\documentclass[lettersize,journal]{IEEEtran}
\usepackage{amsmath,amsfonts}
\usepackage{algorithmic}
\usepackage{algorithm}
\usepackage{array}
\usepackage[caption=false,font=normalsize,labelfont=sf,textfont=sf]{subfig}
\usepackage{textcomp}
\usepackage{stfloats}
\usepackage{url}
\usepackage{verbatim}
\usepackage{graphicx}
\usepackage{cite}
\usepackage{color}
\usepackage{algorithm}
\usepackage{algorithmic}

\usepackage{multicol}
\usepackage{multirow}
\usepackage{makecell}
\usepackage{color}
\usepackage{colortbl}  
\definecolor{gray}{RGB}{150,150,150}
\definecolor{code}{RGB}{255,51,255}

\usepackage{amsmath}
\usepackage{amssymb}
\usepackage{verbatim}
\usepackage{bbm}

\newcommand{\tabincell}[2]{\begin{tabular}{@{}#1@{}}#2\end{tabular}}

\begin{document}

\title{Self-Supervised Modality-Aware Multiple Granularity Pre-Training for RGB-Infrared Person Re-Identification}

\author{Lin Wan, Qianyan Jing, Zongyuan Sun, Chuang Zhang, Zhihang Li, and Yehansen Chen
\thanks{Lin Wan, Yehansen Chen, Qianyan Jing, Zongyuan Sun are with the School of Computer Science, China University of Geosciences, Wuhan 430078, China (e-mail: wanlin@cug.edu.cn; chenyehansen@gmail.com; jingqianyan@cug.edu.cn; sunzongyuan@cug.edu.cn) (Corresponding authors: Zhihang Li and Yehansen Chen)}
\thanks{Chuang Zhang is with the School of Computer Science and Engineering, Nanjing University of Science and Technology, Nanjing 210094, China (e-mail: c.zhang@njust.edu.cn)}
\thanks{Zhihang Li is with the University of Chinese Academy of Sciences, Beijing 100190, China (e-mail: lizhihang.cas@gmail.com)}
}

\markboth{Journal of \LaTeX\ Class Files,~Vol.~14, No.~8, August~2021}%
{Shell \MakeLowercase{\textit{et al.}}: A Sample Article Using IEEEtran.cls for IEEE Journals}


\maketitle

\begin{abstract}
RGB-Infrared person re-identification (RGB-IR ReID) aims to associate people across disjoint RGB and IR camera views. Currently, state-of-the-art performance of RGB-IR ReID is not as impressive as that of conventional ReID. Much of that is due to the notorious modality bias training issue brought by the single-modality ImageNet pre-training, which might yield RGB-biased representations that severely hinder the cross-modality image retrieval. This paper makes first attempt to tackle the task from a pre-training perspective. We propose a self-supervised pre-training solution, named Modality-Aware Multiple Granularity Learning (MMGL), which directly trains models from scratch only on multi-modal ReID datasets, but achieving competitive results against ImageNet pre-training, without using any external data or sophisticated tuning tricks. First, we develop a simple-but-effective `permutation recovery' pretext task that globally maps shuffled RGB-IR images into a shared latent permutation space, providing modality-invariant global representations for downstream ReID tasks. Second, we present a part-aware cycle-contrastive (PCC) learning strategy that utilizes cross-modality cycle-consistency to maximize agreement between semantically similar RGB-IR image patches. This enables contrastive learning for the unpaired multi-modal scenarios, further improving the discriminability of local features without laborious instance augmentation. Based on these designs, MMGL effectively alleviates the modality bias training problem. Extensive experiments demonstrate that it learns better representations ($+$8.03\% Rank-1 accuracy) with faster training speed (converge only in few hours) and higher data efficiency ($<$5\% data size) than ImageNet pre-training. The results also suggest it generalizes well to various existing models, losses and has promising transferability across datasets. The code will be released at \textcolor{code}{https://github.com/hansonchen1996/MMGL}. 
\end{abstract}


\section{Introduction}

\IEEEPARstart{I}t is not a secret that today's success of person re-identification (ReID) is mainly owed to the accumulation of visible images \cite{fu2021cm}. Deep models trained with massive RGB data have spawned beyond-human performance over various benchmarks \cite{zheng2015scalable}. Despite of these advances, it is worth noting that visible cameras are sensitive to illumination variations, often failing to capture valid visual information in low lighting conditions (\textit{e.g.}, at night). Fortunately, owing to the inherent illumination robustness, infrared (IR) images provide effective complement for building a 24-hour ReID system. This has attracted considerable attention to pedestrian retrieval across RGB and IR sensing modalities, \textit{i.e.}, RGB-IR ReID \cite{wu2017rgb}.

\begin{figure}[t]
\begin{center}
   \includegraphics[width=1\linewidth]{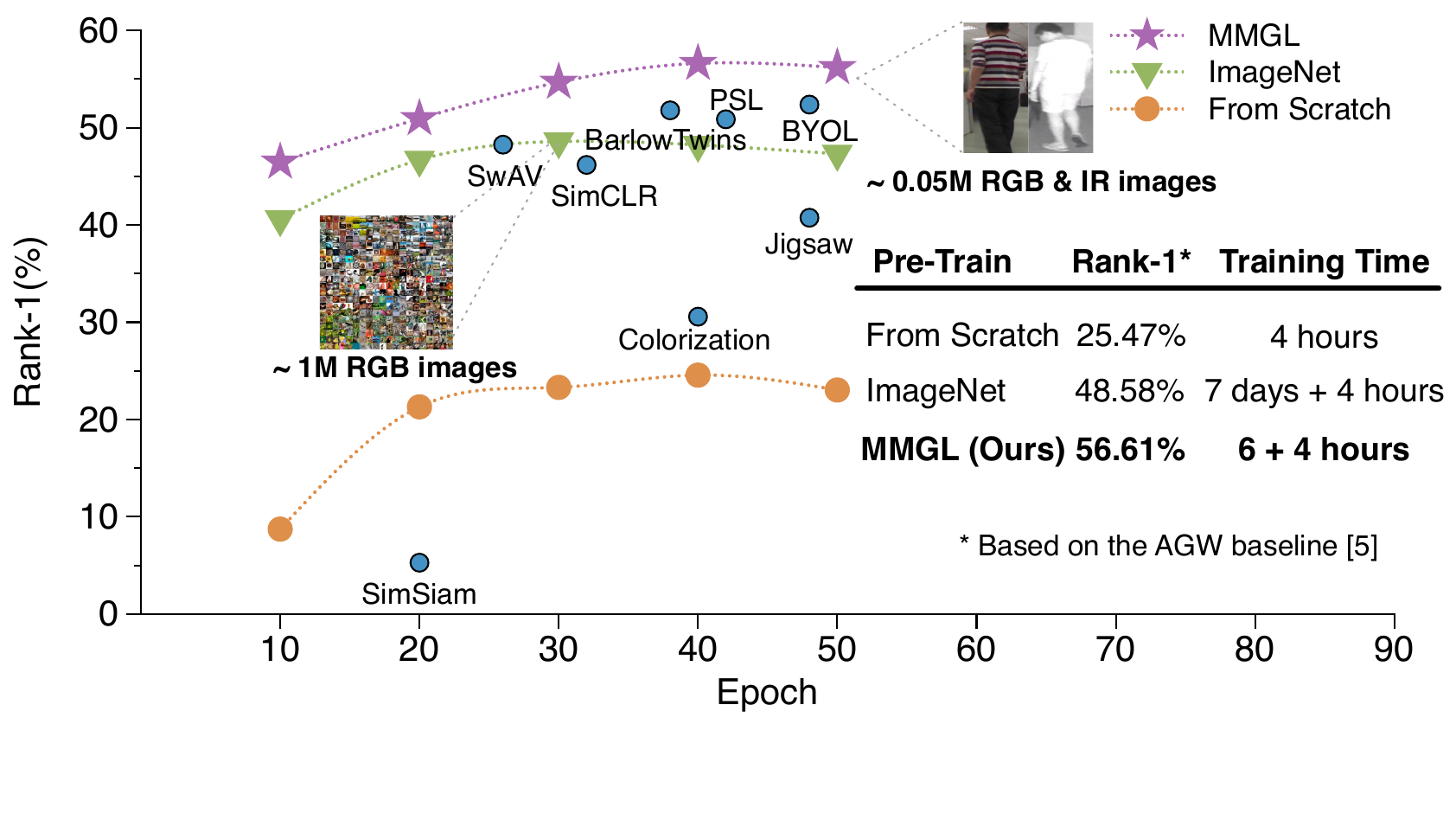}
\end{center}
\vspace{-1.2cm}
   \caption{Comparison of Modality-Aware Multiple Granularity (MMGL) and ImageNet pre-training. MMGL boosts ReID performance with solid data efficiency and also outperforms other self-supervised methods. Times are measured on an Nvidia 2080Ti GPU.}
\label{fig:problem}
\vspace{-0.3cm}
\end{figure}

Facing the lack of large-scale public datasets, pre-training backbones on ImageNet \cite{deng2009imagenet}, and then fine-tuning on target datasets has become a de-facto paradigm for ReID tasks \cite{ye2021deep}. Recently, starting from ImageNet pre-trained checkpoints, Ye \textit{et al.} \cite{ye2021deep} present a powerful AGW baseline for different ReID tasks. Although achieving 95\% Rank-1 accuracy for single-modality ReID (as shown in Fig. \ref{fig:problem}), it obtains only small performance gains on RGB-IR ReID datasets (Rank-1 increases only from 25\% to 48\%). This is because, in RGB-IR ReID, the domain shift between ImageNet and target multi-modality datasets is so large that the marginal benefit from large-scale data \cite{he2019rethinking} is partly counteracted by the \textit{modality bias training} issue \cite{huang2021alleviating}. That is, huge amounts of pre-learned RGB information could overwhelm the `scarce' IR spectrum during fine-tuning, leading to biased representations. To our best knowledge, Huang \textit{et al.} \cite{huang2021alleviating} is the first to investigate the \textit{modality bias training} issue in RGB-IR ReID. Despite exhibiting considerable performance, their solution requires a task-specific image generator and loss functions for the fine-tuning stage, which might degrade the generalization ability of learned representations. Moreover, they still adopt ImageNet-pretrained models, costing dearly in both training time and computational resources while not solving the problem fundamentally. This motivates us to think: \textit{Is ImageNet pre-training the only start point for cross-modality image retrieval?}

One intuitive alternative might be `without pre-training', but `training from scratch' is not a feasible way owing to the accompanying \textit{over-fitting} problem. As shown by the brown line in Fig. \ref{fig:problem}, directly training models from scratch on existing RGB-IR ReID datasets has not achieved competitive results as \cite{he2019rethinking}. To mitigate over-fitting, Fu 
\textit{et al.}\cite{fu2021unsupervised} perform unsupervised pre-training on a huge LUPerson dataset, gaining significant performance boost on mainstream ReID tasks. However, such improvement still heavily relies on \textit{massive} human-collected data ($\sim$4M). Besides, this dataset also only contains RGB images, which is unable to alleviate the above-mentioned \textit{modality bias training} issue.

This paper questions \textit{modality bias training} and \textit{expensive pre-training} in current RGB-IR ReID by exploring a novel regime: we report that competitive cross-modality ReID accuracy can be achievable when directly pre-training on \textit{target RGB-IR datasets, without additional data and manual labels}. More interestingly, without bells and whistles, it even \textit{surpasses} ImageNet supervised pre-training by a large margin with $<$5\% data size (0.05M vs. 1M) and 20$\times$ training speed (see Fig. \ref{fig:problem} for details). The secret to our success mainly lies in two simple but effective designs. \textbf{First}, we develop a \textit{permutation recovery} pretext task, training an encoder end-to-end to recover the original order of randomly shuffled person images. By globally mapping RGB-IR image pairs into the same permutations, a shared permutation latent space will be learned to narrow the distribution gap between RGB and IR pixels, yielding modality-invariant representations. \textbf{Second}, based on the shuffled image patches, we further advance a part-aware cycle-contrastive learning strategy to capture fine-grained visual cues at the local granularity. Given a specific patch, we treat it as a query to derive two attentive representations within/across modality using soft nearest neighbor retrieval. The retrieved patches form a cycle between modalities, acting as a positive pair for contrastive learning \cite{hjelm2018learning}. Employing such cross-modality cycle-consistency achieves contrastive learning with unpaired multi-modal images, effectively improving the local discriminability of learned representations. Compared with prior work \cite{fu2021unsupervised}, our formulation uses natural cross-modality correlations to avoid laborious data augmentation, allowing efficient RGB-IR ReID pre-training.

When directly pre-training a simple baseline \cite{ye2021deep} from scratch, our MMGL approach achieves \textbf{8.03\% absolute improvement} over its ImageNet counterpart (Fig. \ref{fig:problem}, Purple Line). Extensive experiments demonstrate that such comparability holds when applying MMGL to various state-of-the-art ReID methods. Furthermore, the pre-trained models also show better generalization capability in cross-dataset settings. Our contributions are three-fold:

\begin{itemize}
    \item We are the first to explore pre-training solutions for RGB-IR ReID and pioneer a non-ImageNet-powered paradigm which implements self-supervised pre-training directly on target datasets, effectively solving the \textit{modality bias training} issue with promising data efficiency and training speed, achieving competitive task performance against ImageNet pre-training. 
    \item We propose a part-aware cycle-contrastive learning strategy to achieve contrastive learning for the unpaired multi-modal ReID scenarios, which further increases the discriminability of cycle-consistent RGB-IR patches, significantly improving the task performance.
    \item We conduct extensive experiments to show that our self-supervised pre-training approach gains promising RGB-IR ReID performance, generalizes well to various state-of-the-art ReID models, losses and has promising transferability across datasets.
\end{itemize}

The rest of the paper is organized as follows. Section \ref{sec:related_work} provides a brief review on recent progress of single-modality and multi-modality person ReID, as well as self-supervised learning. Section \ref{sec:method} gives a detailed description of our proposed MMGL pre-trainig paradigm. Extensive experiments on the performance and generalization ability of MMGL are presented in Section \ref{sec:experiments}. Section \ref{sec:discussion} includes an in-depth discussion on the broader impact and limitations of the proposed method. Conclusions are drawn in Section \ref{sec:conclusion}.

\section{Related Work}
\label{sec:related_work}

\subsection{RGB-based Person ReID}
RGB-based ReID aims to match a target pedestrian across disjoint visible camera views at varying places and times \cite{zheng2015scalable}. It is challenging to learn suitable feature representations robust enough to withstand large intra-class variations of illumination \cite{zeng2020illumination}, poses \cite{miao2019pose}, scales \cite{hou2019interaction}, viewpoints \cite{karanam2015person} and background clutter \cite{luo2019spectral}. Earlier studies often utilize handcrafted descriptors (\textit{e.g.,} HOG \cite{dalal2005histograms} and LOMO \cite{liao2015person}) to extract low-level features such as color, shape, or texture for person retrieval. However, It is so hard to manually design optimal visual features with stronger discriminability and better stability \cite{ding2020multi, yan2020learning}. Nowadays, deep learning methods show powerful capacity of automatically extracting features from large-scale image datasets and have achieved state-of-the-art results for RGB-based person ReID tasks \cite{miao2019pose, hou2019interaction, ding2020multi, yan2020learning, wang2019color}. Building on various sophisticated CNN architectures, deep ReID models are doing exceptionally well on visual matching by learning discriminatory cross-camera feature representations and optimal distance metrics in an end-to-end manner \cite{ye2021deep, chen2017person, chen2017beyond}. To learn more discriminative features, part information and contextual information is also widely exploited in recent works \cite{wang2020high, kalayeh2018human, miao2019pose, shen2018person, luo2019spectral}. Although these approaches have reported encouraging performance, it is infeasible to directly employ them in real-world surveillance scenarios with poor lighting environments \cite{ye2020visible, ye2020cross}.

\subsection{RGB-IR Person ReID}
RGB-IR Person ReID focuses efforts in tackling pixel and feature misalignment issues between modalities \cite{ye2021deep}. Currently, existing methods can be roughly divided into two lines: image synthesis and shared feature learning. Image synthesis methods usually adopt generative adversarial networks (GAN) \cite{goodfellow2014generative} to minimize pixel-level difference across modality by synthesizing IR/RGB counterparts for RGB/IR images. Following this vein, \cite{wang2019rgb} firstly propose to combine pixel and feature alignment to learn modality-invariant and discriminative representations. Several studies also apply cross-modality paired image synthesis \cite{wang2020cross}, feature disentanglement \cite{choi2020hi}, and intermediate modality generation \cite{li2020infrared} to enhance the quality of generated images. However, it is ill-posed for GAN-based methods to recover color appearances for IR images, resulting in the deficiency of identity information in fake RGB images \cite{ye2019bi}. Shared feature learning attempts to discover a common feature subspace to align modality distributions. \cite{wu2017rgb} first release a multimodal ReID dataset (SYSU-MM01) and present a deep zero-padding network for RGB-IR ReID. \cite{ye2021channel} propose a two-stream network to extract shared features. Recently, various feature selection approaches are designed to enhance representation discriminability, including graph neural networks \cite{ye2020dynamic} and automated feature search \cite{chen2021neural}. However, they all adopt ImageNet pre-training as an outset, suffering from the modality bias brought by the single-modality pre-training.

\subsection{Self-Supervised Learning}
Self-Supervised Learning (SSL) is currently the fastest-growing branch of unsupervised learning, aiming to learn `universal' representations by solving a pre-designed \textit{pretext} task using data itself as supervisory signals \cite{jing2020self}. Over the last decade, self-supervised learning has witnessed a wide range of pretext task designs based on data generation \cite{zhang2016colorful}, spatial or temporal contexts \cite{noroozi2016unsupervised}, and multi-modal correspondence \cite{zou2018df}. For various types of self-supervised pretext tasks, contrastive supervision aims to maximize agreement between different data views to forge a discriminative feature subspace \cite{caron2020unsupervised,grill2020bootstrap,zbontar2021barlow,chen2021jigsaw,chen2021exploring}, which has been proven effective in learning high quality representations for downstream tasks \cite{chen2020simple,he2020momentum}. Recent advances on contrastive learning focus on learning contrastive correspondence with delicately designed positive/negative pair generation strategies. For instance, MoCo \cite{he2020momentum} constructs negative pairs with a dynamic cross-batch queue; BYOL \cite{grill2020bootstrap} uses a slow-moving average of the online network to replace negative samples; and HCSC \cite{guo2022hcsc} leverages the hierarchical semantic structures of data to retrieve both positive and negative pairs. Nevertheless, in multi-modal ReID scenarios, contrastive correspondence is difficult to be mined due to the unpaired nature of heterogeneous images. In this paper, we propose to exploit cycle-consistency \cite{zhu2017unpaired} between body parts to implement contrastive learning for RGB-IR ReID scenarios.

\section{Methodology}
\label{sec:method}

\begin{figure*}[t]
\begin{center}
   \includegraphics[width=1\linewidth]{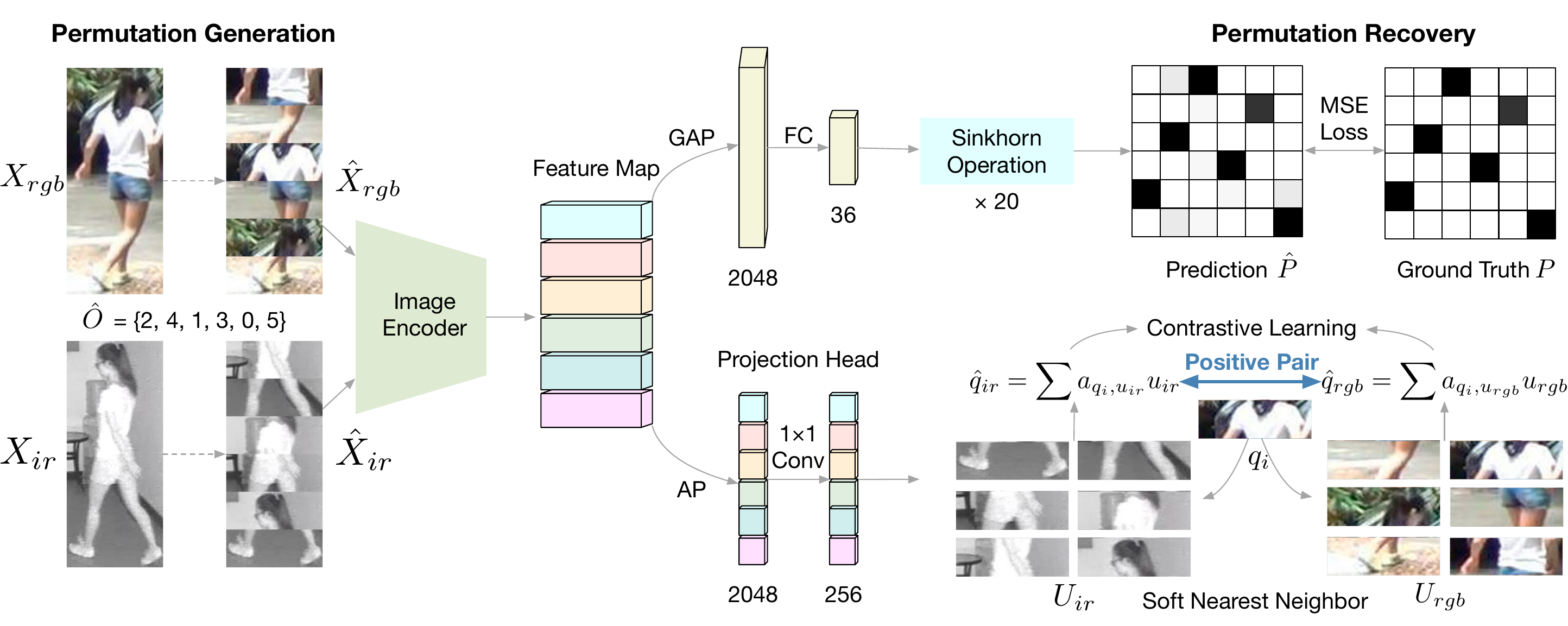}
\end{center}
   \caption{An overview of the proposed  Modality-Aware Multiple Granularity Learning (MMGL) pipeline. Generally, MMGL consists of a permutation recovery branch and a cycle-contrast branch at global and local granularities respectively. The former aims to learn a shared permutation latent space by recovering the original order of each cross-modality shuffled image pair. The latter seeks to enhance patch discriminative power by maximizing the agreement between patch representations derived with cross-modality cycle-consistency.}
\label{fig:pipeline}
\end{figure*}

\subsection{Problem Formulation}
\label{sec:formulation}

RGB-IR person ReID can be formulated as a cross-modality image retrieval problem, where query and gallery images are from different modalities. This section takes the \textit{Infrared to Visible} setting for the problem formulation. Given a query image $x^{ir}_{q}$ of a person $y^{ir}_{q}$, RGB-IR ReID seeks to return a ranking list $\{(x^{rgb}_{1}, y^{rgb}_{1}), (x^{rgb}_{2}, y^{rgb}_{2}), \dots, (x^{rgb}_{n}, y^{rgb}_{n})\}$ on the visible gallery set, where $x^{rgb}_{i}$ indicates a visible gallery image, $
y_{i}$ denotes its corresponding identity, and $i$ is the sorted image index. In this ranking list, all $x^{rgb}_{i}$ with $y^{rgb}_{i}=y^{ir}_{q}$ should be ranked to the top. Following the Normalized Discounted Cumulative Gain (NDCG) \cite{jarvelin2017ir}, it requires to optimize the following objective function,
\begin{equation}
\label{eq:retrieval_objective}
\begin{aligned}
    \max&\!\sum_{k=1}^{n}\!\frac{\mathbbm{1}(y^{ir}_q, y^{rgb}_k)}{k}\!\sum_{i=1}^{k}\!\mathbbm{1}(y^{ir}_{q}, y^{rgb}_{i})\!\quad\!,  \\
    &\mathbbm{1}(y^{ir}_{q}, y^{rgb}_{i})=
\begin{cases}
      1,& \!\text{$y^{ir}_{q} = y^{rgb}_{i}$}\\
      0,& \!\text{$y^{ir}_{q} \neq y^{rgb}_{i}$}
\end{cases}
\end{aligned}
\end{equation}
where $y^{ir}_{q}$ denotes the identity information of the infrared query image, and $n$ is the size of the visible gallery set.

Prior work \cite{ye2020dynamic} suggests that solving Eq. (\ref{eq:retrieval_objective}) requires the learned representations being both modality-invariant and discriminative. This motivates us to propose two novel designs in our MMGL pre-training paradigm: \textbf{(a)} \textit{Permutation Recovery} pretext task that maps randomly shuffled RGB-IR image pairs into a shared permutation latent space for global invariant learning. \textbf{(b)} \textit{Part-aware Cycle-Contrastive Learning} strategy that maximizes agreement between cycle-consistent RGB-IR image patches to improve local discriminability. Fig. \ref{fig:pipeline} illustrates our main idea, which will be introduced next. 

\subsection{Cross-Modality Permutation Recovery}
As shown in Fig. \ref{fig:pipeline}, given a ranking vector $\hat{O}$ that represents a randomly shuffled image patch sequence, the permutation recovery task aims to learn to reconstruct its original counterpart $O$ with a \textit{permutation matrix} $P$ \cite{mena2018learning}. Mathematically, $P$ belongs to the set of 0-1 doubly stochastic matrices, wherein each non-zero element in the $i$-th row and $j$-th column suggests that the current $i$-th patch should be assigned to the $j$-th place of the sequence. This leads to a regression problem $O=P_{\Theta, \hat{O}}^{-1}\hat{O}$ for which we seek to derive $P$ with network parameters $\Theta$. However, the discrete nature of this problem may pose great challenges to the model optimization. Here, we present how to approximate this task solving process in the context of backpropagation.

\textbf{Permutation Generation.} We introduce a modality-shared shuffling operator $G(X_{rgb},X_{ir},\hat{O})$ that transforms randomly composed cross-modality image pairs $\{X_{rgb},X_{ir}\}$ to their shuffled counterparts $\{\hat{X}_{rgb},\hat{X}_{ir}\}$ (see Fig. \ref{fig:pipeline}, left). Suppose the shuffled image contains $N$ patches, $\hat{O}$ is given by a random permutation of an array $[1,\cdot\cdot\cdot,N]$ sampled from the uniform distribution, which denotes the mapping from original patches to their randomly shuffled ones. For each image with $H$ height and $W$ width, the shuffling generates a rearranged sequence of image patches (with size of $H/N\times W$) by choosing the $i$-th patch from $\hat{O}_{i}$ position of the original sequence. Note that $\hat{O}$ is shared within each cross-modality image pair, which maps both modality images into a common permutation subspace. This is beneficial to learn modality-invariant representations.

\textbf{Permutation Recovery.} We map shuffled images to corresponding affinity matrices $\hat{P}\in\mathbb{R}^{N\times N}$ to recover their original orders. Such mappings can be fitted by a encoder $\mathcal{F}(\hat{X}_{rgb}, \hat{X}_{ir}, \Theta)$ with parameters $\Theta$ that transforms each image into a $N^{2}$-dim feature representation. Specifically, for each $\{\hat{X}_{rgb}, \hat{X}_{ir}\}$, $\mathcal{F}$ learns two global representations $f_{rgb}$ and $f_{ir}$. We then reduce their dimensions to $N^{2}$ using a shared fully-connected layer $\mathcal{G}$, \textit{i.e.,} $\hat{f}_{rgb}, \hat{f}_{ir}=\mathcal{G}(f_{rgb}, f_{ir})$. It is worth noting that the selection of $\mathcal{F}$ should be consonant with downstream supervised models. We transform $\hat{f}$ into a $N\times N$ matrix $\hat{P}$, in which each row and column can be regarded as a logit vector that denotes the possibility of a patch belonging to a serial position (\textit{i.e.,} a category).

Nevertheless, it is tough to directly fit permutation matrix $P$ with the learned $\hat{P}$, because each patch is discrete, making the approximation process non-differentiable. To this end, we introduce the Gumbel-Sinkhorn operator \cite{mena2018learning} to relax $\hat{f}$ to the continuous domain so as to fit a categorical distribution. The Sinkhorn operator is defined as:
\begin{equation}
\begin{aligned}
\operatorname{Sinkhorn}^{0}(\hat{P}) &=\exp (\hat{P}), \\
\operatorname{Sinkhorn}^{l}(\hat{P}) &=\mathcal{T}_{c}\left(\mathcal{T}_{r}\left(\operatorname{Sinkhorn}^{l-1}(\hat{P})\right)\right), \\
\operatorname{Sinkhorn}(\hat{P}) &=\lim _{l \rightarrow \infty} \operatorname{Sinkhorn}^{l}(\hat{P}),
\end{aligned}
\end{equation}
where $\mathcal{T}_{r}(\hat{P})=\hat{P} \oslash\left(\hat{P} \mathbf{1}_{N} \mathbf{1}_{N}^{\top}\right)$, $\mathcal{T}_{c}(\hat{P})=\hat{P} \oslash\left(\mathbf{1}_{N} \mathbf{1}_{N}^{\top} \hat{P}\right)$ indicates the row and column-wise normalization of a matrix respectively, $\oslash$ means element-wise division, $\mathbf{1}_{N}$ is a column vector of ones, $l$ is the number of iterations.

Based on the Sinkhorn operator, we can reparameterize the hard choice of the permutation matrix with Gumbel-Softmax distribution \cite{jang2016categorical}:
\begin{equation}
\operatorname{G-Sinkhorn}(\hat{P} / \tau)={\operatorname{softmax} }((\operatorname{trace}(P^{\top}\hat{P})+\gamma)/\tau),
\end{equation}
where $\gamma$ denotes random noises sampled from a Gumbel distribution, $\tau$ is the temperature hyperparameter.

After relaxing the learned affinity matrix $\hat{P}$, we could gradually approach the ground truth $P$ via backpropagation, aiming to minimize the following permutation reconstruction error:
\begin{equation}
\mathcal{L}_{p}=\sum_{i=1}^{M}\left\|O_{i}-\hat{P}^{-1} \hat{O}_{i}\right\|^{2},
\end{equation}
where $M$ is the size of a mini-batch, $O$ and $\hat{O}$ is the original/shuffled ranking vector, respectively.

\textbf{Comparisons with Existing Work.} Solving jigsaw serves as the most relevant pretext task to our permutation recovery and intuitively forms a more diverse feature space. However, we empirically find that jigsaw task presents inferior performance (see Table \ref{Table:SSL}), which is quite counter-intuitive. One possible reason is that the pose variations in person images may incur severe spatial misalignment in horizontal direction, making patches semantically meaningless. Besides, current jigsaw-oriented pre-training \cite{noroozi2016unsupervised,li2021progressive} mainly concentrates on learning patch-wise representations, while overlooking the contexts between patches. One solution to this issue might be introducing patch overlap \cite{chen2021jigsaw}, but it will inevitably bring additional computational costs. Our permutation recovery directly takes full images as inputs to encode contexts, proven to be more powerful and efficient (like YOLO \cite{redmon2016you} vs. R-CNN \cite{girshick2014rich}).

\vspace{-0.3cm}
\subsection{Part-Aware Cycle-Contrastive Learning}
With cross-modality permutation recovery, the pre-trained model is able to learn modality-invariant biometrics (\textit{e.g.,} shape and texture) in favour of modality alignment. However, it may collapse to \textit{low-loss} solutions that hinder it learning desired representations, e.g., directly utilizing boundary patterns and textures continuing across patches to solve the task. Such `\textit{shortcuts}' will suppress the intra-class compactness, leading to fuzzy decision boundaries for identity recognition.

Recent advances on contrastive learning show its promising capability in learning discriminative representations \cite{chen2020simple,he2020momentum}. With delicately designed `two-crop augmentation' strategies, it maximums the agreement between different views of the same image to achieve better intra-class compactness and inter-class discriminability. Nonetheless, due to the unpaired nature of heterogeneous images in RGB-IR ReID task, it is infeasible to directly apply off-the-shelf contrastive learning pipelines to the multi-modal scenario. As most augmentations are intra-modality, they also cannot reflect cross-modality correlations of identity semantics.

Here, we propose a part-aware cycle-contrastive (PCC) constraint that uses cross-modality cycle consistency to conduct contrastive learning at the local granularity. Rather than two-crop augmentation, PCC adopts the soft-nearest neighbor retrieval to generate positive patch pairs. Specifically, for each image patch, we utilize a \textit{PCB-style} projection head \cite{sun2019learning} to map it into a normalized 256-dim representation (see Fig. \ref{fig:pipeline}, lower branch). After projection, we introduce a forward-backward nearest neighbor process to capture cross-modality cycle-consistency. Given a query representation $q_{i}$, we first derive its soft nearest neighbor $\hat{q}$ from the universal representation set $U$ of counterpart modality. Then, we compute the soft nearest neighbor of $\hat{q}$ backwards within the patch set of the same image. The cycle-consistency is satisfied when the two retrieved soft nearest neighbors are similar. The retrieval process is defined as:
\begin{equation}
\hat{q}_{i}=\sum_{u \in U} \alpha_{q_{i}, u} u,\quad\alpha_{q_{i}, u}=\frac{\exp \left(\operatorname{sim}\left(q_{i}, u\right) / \tau\right)}{\sum_{u^{\prime} \in U} \exp \left(\operatorname{sim}\left(q_{i}, u^{\prime}\right) / \tau\right)},
\end{equation}
where $\tau$ is the temperature and $\operatorname{sim}$ is the cosine similarity.

After the similarity-based forward-backward retrieval, we would obtain a cross-modality patch pair $\{\hat{q}_{ir},\hat{q}_{rgb}\}$ with similar semantics and two sets of retrieved representations $\{V_{ir}, V_{rgb}\}$. We regard $\{\hat{q}_{ir},\hat{q}_{rgb}\}$ as a positive pair while $\{V_{ir}, V_{rgb}\}$ as negative sets for contrastive learning:
\begin{equation}
\label{eq:PCC}
\mathcal{L}_{\text {PCC}}=-\log \frac{\exp \left(\operatorname{sim}\left(\hat{q}_{ir}, \hat{q}_{rgb}\right) / \tau\right)}{\sum_{u \in\left\{V_{ir}, V_{rgb}\right\}} \exp \left(\operatorname{sim}\left(\hat{q}_{i}, u\right) / \tau\right)}.
\end{equation}

By pulling together $\{\hat{q}_{ir},\hat{q}_{rgb}\}$ whilst pushing away all negative pairs, the deep network will learn to discover modality correspondence across semantic-similar body partitions. This explicit supervision facilitates the fine-grained alignment of heterogeneous images, helping to transfer better modality invariance and discriminative power to downstream cross-modality ReID models. 

Since the encoder is pre-trained with random initialization, PCC might generate random positive pairs in early training, and that will degrade the discriminative power of learned representations. To avoid this, we need a penalization term that guides PCC to retrieve semantic-similar patches for positive pair generation. Here, we further impose a cross-entropy loss $\mathcal{L}_\text{cls}$ on part representations to predict the permutation label of each patch. With this supervision, patches of the same body part are encouraged to be closer in the embedding space and more likely to be regarded as positive pairs.

The overall learning objective of MMGL is formulated as:
\begin{equation}
\label{eq:all_loss}
    \mathcal{L}_{\text{MMGL}}=\mathcal{L}_{p}+\mathcal{L}_{\text {cls}}+\lambda\mathcal{L}_{\text {PCC}},
\end{equation}
where $\lambda$ is a trade-off factor to balance each objective.

\textbf{Comparisons with Existing Work.} Following the general paradigm of contrastive learning, PCC essentially solves an instance discrimination pretext task by maximizing agreement between different views of the same person image. But unlike existing methods, for multi-modal scenarios, we introduce two novel designs: \textbf{1)} creating views without the need of data augmentation, and \textbf{2)} cross-modality cycle-consistency for positive pair generation.

Existing contrastive learning frameworks \cite{chen2020simple,he2020momentum,chen2021exploring,caron2020unsupervised,grill2020bootstrap,zbontar2021barlow} largely depend on `two-crop augmentation' to generate positive pairs. Recently, \cite{tian2020makes} reveals that the performance of `two-crop augmentation' is sensitive to specific augmentation workflows. In contrast to these methods, PCC explicitly exploits the natural `views' for RGB-IR ReID, i.e., RGB and IR modalities, to decouple data augmentation from contrastive learning, which simultaneously improves the pre-training performance and efficiency (see Table \ref{Table:SSL} for empirical results).

On the other hand, inspired by CycleGAN \cite{zhu2017unpaired}, PCC adopts cross-modality cycle-consistency to retrieve semantically consistent pairs, which enables contrastive learning among unpaired image patches and that further improves local discriminability in RGB-IR ReID.

\subsection{Supervised Fine-Tuning for RGB-IR ReID}
Since MMGL is a loss function-based algorithm, it is agnostic to the choice of encoder $\mathcal{F}$ and can serve as a universal front-end in tandem with any RGB-IR ReID models. In the fine-tuning stage, we transfer the MMGL pre-trained backbone to downstream models and perform supervised learning for cross-modality image retrieval. Following \cite{ye2020dynamic}, we optimize the identity cross-entropy loss $\mathcal{L}_{\text{id}}$, triplet loss $\mathcal{L}_\text{triplet}$ \cite{hermans2017defense}, and method-specific loss $\mathcal{L}_\text{specific}$ for different approaches, i.e.,
\begin{equation}
\label{loss:ReID}
    \mathcal{L}_{\text{ReID}}=\mathcal{L}_{\text{id}}+\mathcal{L}_{\text{triplet}}+\mathcal{L}_\text{specific}.
\end{equation}

\section{Experiments}
\label{sec:experiments}

In this section, we first introduce the experimental databases, evaluation protocols, and implementation details. Then, the proposed MMGL pre-training paradigm is equipped on several state-of-the-art RGB-IR ReID methods to verify its effectiveness in learning modality-invariant and discriminative representations. Thereafter, we evaluate the cross-dataset generalization ability of MMGL through transfer learning. Furthermore, we conduct ablation studies to investigate the contribution of each component and analyze some visualization results. Finally, we make in-depth discussions on the board impact and limitations of our pre-training algorithm.

\subsection{Datasets and Experimental Settings}

\textbf{The SYSU-MM01 dataset} \cite{wu2017rgb}.  This is a large-scale benchmark dataset for RGB-IR ReID, including 287,628 RGB and 15,792 IR images captured by four visible and two infrared cameras. All images are randomly obtained without a one-to-one correlation between modalities. The original database is randomly split into the training set and the testing set, comprising 395 and 96 person identities respectively. In our experiments, we adopt both \textit{all-search} and \textit{indoor-search} modes to evaluate the model performance. The \textit{all-search} gallery set involves images captured by all four visible cameras, while the \textit{indoor-search} gallery set only contains images captured by two indoor visible cameras. To evaluate the impact of different gallery sizes, we further adopt \textit{single-shot} (1 RGB image per identity) and \textit{multi-shot} (10 RGB images per identity) settings \cite{wu2017rgb} for both search modes. For all settings, the query set involves 3,803 IR images of 96 identities.

\textbf{The RegDB dataset} \cite{nguyen2017person}. This dataset is collected by a dual camera acquisition system, including one visible and one thermal camera. The two cameras are bound together to capture paired RGB and IR images of 412 identities (206 identities for training and the others for testing). Each identity has 10 thermal and 10 visible images. Following \cite{wang2020cross}, we evaluate both \textit{visible-to-infrared} and \textit{infrared-to-visible} modes. The former takes thermal images as gallery set and visible images as query set, while the latter is the opposite.

\textbf{Evaluation protocols.} We follow the general evaluation protocol \cite{wu2017rgb, ye2020dynamic} for RGB-IR person ReID. Results of
SYSU-MM01 are based on an average of 10 random splits of gallery set and query set. The performance on RegDB is averaged over ten trials on random training-testing splits. The standard cumulated matching characteristics (CMC) curve and mean average precision (mAP) are used to evaluate the retrieval performance.

\subsection{Implementation Details}
We implement MMGL with PyTorch on an Nvidia 2080Ti GPU. All images are resized to 288$\times$144. A SGD optimizer with 0.9 momentum and 0.0005 weight decay is used for optimization. We set initial learning rate to 0.1 with linear warm-up \cite{ye2020dynamic} and decay it with the cosine decay schedule without restarts, training for a total of 100 epochs.

\textbf{Pre-Training Stage.} For pre-training, we randomly sample 48 RGB and 48 IR images without labels for each training batch. Each image is split equally into 6 horizontal stripes. Conventional random cropping with zero-padding, horizontal flipping, AugMix \cite{hendrycks2019augmix} and random erasing \cite{zhong2020random} are chosen for data augmentation. Following \cite{mena2018learning}, we use $l=20$ for Sinkhorn operator iterations. For each sample, we generate 10 reconstructions using Gumbel perturbations. Following \cite{he2020momentum}, we set $\tau$ to 0.07. And $\lambda$ is empirically set to 0.2.

\textbf{Fine-Tuning Stage.} For fine-tuning, we drop the \textit{PCB-style} projection head and initialize the model with MMGL checkpoints, while \textit{leaving other default settings (\textit{e.g.,} hyperparameters and data augmentation strategies) unchanged}. Note that these settings are not customized for MMGL, tuning them likely leads to better results. We leave it to future work.

\textbf{Backbone Architectures.} In our experiments, we consider two baseline architectures, i.e., \textit{one-stream} and \textit{two-stream} networks. Both of them adopt ResNet-50 \cite{he2016deep} as the backbone.

Architecture details of backbone networks are shown in Table \ref{table:structure}. The \textit{one-stream} network shares all blocks for different modalities. And the \textit{two-stream} one makes the first convolutional block independent for each modality while sharing the other blocks for global feature representation learning. All images are resized to 288 $\times$ 144 as the inputs. The stride of the last convolutional block is set to 1 so as to obtain fine-grained feature maps \cite{luo2019bag}. The other hyper-parameters are following \cite{he2016deep} without tuning.

\begin{table}[t]\scriptsize
\caption{Architecture details of the used backbones.}
\label{table:structure}
\begin{center}
\begin{tabular}{c|c|c|c}
\hline
Layer name             & Output size            & 50-layer & Type \\ \hline
\multirow{2}{*}{Conv1}                  & \multirow{2}{*}{144$\times$72}                 & 7$\times$7, 64, stride 2               &  \multirow{2}{*}{Shared/Specific} \\ \cline{3-3} &    &3$\times$3 max pool, stride 2    &                \\ \hline
Conv2 & 72$\times$36 & \tabincell{c}{\\ $\begin{bmatrix} 1\times1, 64 \\  3\times3, 64 \\ 1\times1, 256 \end{bmatrix} \times 3$ \\ \quad} &Shared   \\ \hline 
Conv3                  & 36$\times$18                  &  \tabincell{c}{\\ $\begin{bmatrix} 1\times1, 128 \\  3\times3, 128 \\ 1\times1, 512 \end{bmatrix} \times 4$  \\ \quad}                              &     Shared                \\ \hline
Conv4                  & 18$\times$9                   &   \tabincell{c}{\\$\begin{bmatrix} 1\times1, 256 \\  3\times3, 256 \\ 1\times1, 1024 \end{bmatrix} \times 6$ \\ \quad}                               &     Shared                \\ \hline
Conv5                  & 18$\times$9                   &   \tabincell{c}{\\$\begin{bmatrix} 1\times1, 512 \\  3\times3, 512 \\ 1\times1, 2048 \end{bmatrix} \times 3$ \\ \quad}                             &     Shared                \\ \hline
\end{tabular}
\end{center}
\end{table}

One merit of MMGL pre-training is that we can directly pre-train the two-stream backbone with high data efficiency. While the ImageNet counterpart can only pre-train one-stream architecture and initialize modality-specific blocks with the same weights. This would aggravate the \textit{modality bias training} issue unfavorable for fine-tuning. Empirically, pre-training two-stream backbones always leads to better RGB-IR ReID performance than one-stream pre-training (see Table \ref{Table:SYSU} \& Table \ref{Table:RegDB} for details).

\renewcommand\arraystretch{1.2}
\begin{table*}[t]

\caption{MMGL with common baselines on SYSU-MM01 with Rank-1, 10, 20 (\%) and mAP (\%) evaluation metrics.}
\label{Table:SYSU}
\resizebox{1\textwidth}{!}{
\begin{tabular}{c|c|c|cccc|cccc|cccccccc}
\multirow{3}{*}{Method} &\multirow{3}{*}{Venue}     & \multirow{3}{*}{Pre-Train} & \multicolumn{8}{c|}{All-Search}                                    & \multicolumn{8}{c}{Indoor-Search}                                          \\ 
                            &                            & & \multicolumn{4}{c|}{Single-Shot} & \multicolumn{4}{c|}{Multi-Shot} & \multicolumn{4}{c|}{Single-Shot}          & \multicolumn{4}{c}{Multi-Shot} \\ 
                            &            &                & r1    & r10    & r20    & mAP    & r1    & r10    & r20    & mAP   & r1 & r10 & r20 & \multicolumn{1}{c|}{mAP} & r1    & r10    & r20   & mAP   \\ \Xhline{1.5pt}
\multirow{3}{*}{One-Stream \cite{ye2021deep}} &\multirow{3}{*}{TPAMI 2021} & \textcolor{gray}{Random Init.}                &\textcolor{gray}{26.29}       &\textcolor{gray}{67.86}        &\textcolor{gray}{81.26}        &\textcolor{gray}{27.28}        &\textcolor{gray}{30.83}       &\textcolor{gray}{73.07}        &\textcolor{gray}{85.53}        &\textcolor{gray}{20.81}       &\textcolor{gray}{26.05}    &\textcolor{gray}{72.15}     &\textcolor{gray}{87.99}     & \multicolumn{1}{c|}{\textcolor{gray}{36.24}}    &\textcolor{gray}{31.39}       &\textcolor{gray}{78.86}        &\textcolor{gray}{91.82}       &\textcolor{gray}{25.96}       \\
                         &   & ImageNet-1k                &43.66      &83.92        &92.23        &43.70        &50.16       &87.10        &94.50        &37.20       &48.93    &89.28     &96.40     & \multicolumn{1}{c|}{57.86}    &56.80       &92.47        &97.11       &49.11       \\
                         &   & MMGL (Ours)                 &\textbf{52.65}       &\textbf{89.31}        &\textbf{95.00}        &\textbf{51.48}        &\textbf{59.99}       &\textbf{92.76}        &\textbf{96.98}        &\textbf{45.03}       &\textbf{55.05}       &\textbf{93.43}        &\textbf{97.02}         &\multicolumn{1}{c|}{\textbf{63.75}}    &\textbf{64.85}       &\textbf{95.53}        &\textbf{98.02}       &\textbf{54.72}      \\ \hline
\multirow{3}{*}{AGW \cite{ye2021deep}}    &\multirow{3}{*}{TPAMI 2021}        & \textcolor{gray}{Random Init.}               &\textcolor{gray}{25.47}       &\textcolor{gray}{66.06}        &\textcolor{gray}{80.11}        &\textcolor{gray}{26.07}        &\textcolor{gray}{32.67}       &\textcolor{gray}{75.81}        &\textcolor{gray}{87.13}        &\textcolor{gray}{22.98}       &\textcolor{gray}{26.38}    &\textcolor{gray}{70.61}     &\textcolor{gray}{86.08}     & \multicolumn{1}{c|}{\textcolor{gray}{35.95}}    &\textcolor{gray}{33.37}       &\textcolor{gray}{80.01}        &\textcolor{gray}{92.15}       &\textcolor{gray}{27.83}       \\
                        &    & ImageNet-1k               &48.58       &87.58        &94.83        &49.37        &54.73       &92.46        &96.72        &42.90       &54.67    &89.56     &95.91     & \multicolumn{1}{c|}{42.90}    &62.13       &93.42        &97.06       &54.56       \\
                        &    & MMGL (Ours)                &\textbf{56.61}       &\textbf{90.23}        &\textbf{95.91}        &\textbf{54.50}        &\textbf{61.41}       &\textbf{93.30}        &\textbf{97.40}        &\textbf{48.27}       &\textbf{62.55}    &\textbf{94.60}     &\textbf{98.02}     & \multicolumn{1}{c|}{\textbf{69.62}}    &\textbf{70.13}       &\textbf{95.55}        &\textbf{98.08}       &\textbf{61.47}  \\
                            \hline
\multirow{3}{*}{DDAG \cite{ye2020dynamic}}   &\multirow{3}{*}{ECCV 2020}        & \textcolor{gray}{Random Init.}               &\textcolor{gray}{fail}       &\textcolor{gray}{fail}        &\textcolor{gray}{fail}        &\textcolor{gray}{fail}        &\textcolor{gray}{fail}       &\textcolor{gray}{fail}        &\textcolor{gray}{fail}        &\textcolor{gray}{fail}       &\textcolor{gray}{fail}    &\textcolor{gray}{fail}     &\textcolor{gray}{fail}     & \multicolumn{1}{c|}{\textcolor{gray}{fail}}    &\textcolor{gray}{fail}       &\textcolor{gray}{fail}        &\textcolor{gray}{fail}       &\textcolor{gray}{fail}       \\
                        &    & ImageNet-1k                &54.75       &90.39        &95.81        &53.02        &\textbf{61.83}       &92.68        &97.49        &\textbf{47.06}       &\textbf{61.02}    &94.06     &\textbf{98.41}     & \multicolumn{1}{c|}{\textbf{67.98}}    &\textbf{69.23}       &95.13        &\textbf{98.31}       &\textbf{59.42}       \\
                        &    & MMGL (Ours)                &\textbf{55.33}       &\textbf{90.91}        &\textbf{96.15}        &\textbf{54.06}        &60.97       &\textbf{92.88}        &\textbf{97.53}        &46.85       &60.71    &93.82     &98.03     & \multicolumn{1}{c|}{\textbf{67.98}}    &67.79       &\textbf{95.70}        &98.20       &59.00 \\
                            \hline
\multirow{3}{*}{NFS \cite{chen2021neural}}    &\multirow{3}{*}{CVPR 2021}        & \textcolor{gray}{Random Init.}               &\textcolor{gray}{30.45}       &\textcolor{gray}{71.83}        &\textcolor{gray}{82.97}        &\textcolor{gray}{31.43}        &\textcolor{gray}{34.18}       &\textcolor{gray}{76.63}        &\textcolor{gray}{87.34}        &\textcolor{gray}{25.01}       &\textcolor{gray}{30.03}    &\textcolor{gray}{75.59}     &\textcolor{gray}{89.57}     & \multicolumn{1}{c|}{\textcolor{gray}{40.38}}    &\textcolor{gray}{34.70}       &\textcolor{gray}{82.66}        &\textcolor{gray}{93.16}       &\textcolor{gray}{30.09}       \\
                        &    & ImageNet-1k                &56.91       &91.34        &96.52        &55.45        &63.51       &94.42        &97.81        &48.56       &62.79    &96.53     &99.07     & \multicolumn{1}{c|}{69.79}    &70.03       &97.70        &99.51       &61.45       \\
                        &    & MMGL (Ours)                &\textbf{61.65}       &\textbf{93.07}        &\textbf{97.89}        &\textbf{59.25}        &\textbf{67.73}       &\textbf{97.00}        &\textbf{98.82}        &\textbf{52.99}       &\textbf{67.34}    &\textbf{98.30}     &\textbf{99.35}     &\multicolumn{1}{c|}{\textbf{72.28}}    &\textbf{75.76}       &\textbf{98.84}        &\textbf{99.66}       &\textbf{65.68} 
\\ \hline

\multirow{3}{*}{CAJ \cite{ye2021channel}}   &\multirow{3}{*}{ICCV 2021}        & \textcolor{gray}{Random Init.}               &\textcolor{gray}{fail}       &\textcolor{gray}{fail}        &\textcolor{gray}{fail}        &\textcolor{gray}{fail}        &\textcolor{gray}{fail}       &\textcolor{gray}{fail}        &\textcolor{gray}{fail}        &\textcolor{gray}{fail}       &\textcolor{gray}{fail}    &\textcolor{gray}{fail}     &\textcolor{gray}{fail}     & \multicolumn{1}{c|}{\textcolor{gray}{fail}}    &\textcolor{gray}{fail}       &\textcolor{gray}{fail}        &\textcolor{gray}{fail}       &\textcolor{gray}{fail}       \\
                        &    & ImageNet-1k                &66.62       &95.44        &98.56        &64.04        &74.07       &96.88        &98.93        &56.75       &71.39    &96.81     &99.31     & \multicolumn{1}{c|}{76.47}    &82.11       &97.95        &99.03       &70.22       \\
                        &    & MMGL (Ours)                & \textbf{68.37}       &\textbf{96.29}        &\textbf{99.03}        &\textbf{66.27}        &\textbf{77.09}       &\textbf{97.99}        &\textbf{99.40}        &\textbf{59.34}       &\textbf{75.25}    &\textbf{98.26}     &\textbf{99.64}     & \multicolumn{1}{c|}{\textbf{79.80}}    &\textbf{85.15}       &\textbf{99.37}        &\textbf{99.88}       &\textbf{73.75} \\ \hline

\multirow{3}{*}{MutualInfo \cite{tian2021farewell}}   &\multirow{3}{*}{CVPR 2021}        & \textcolor{gray}{Random Init.}               &\textcolor{gray}{fail}       &\textcolor{gray}{fail}        &\textcolor{gray}{fail}        &\textcolor{gray}{fail}        &\textcolor{gray}{fail}       &\textcolor{gray}{fail}        &\textcolor{gray}{fail}        &\textcolor{gray}{fail}       &\textcolor{gray}{fail}    &\textcolor{gray}{fail}     &\textcolor{gray}{fail}     & \multicolumn{1}{c|}{\textcolor{gray}{fail}}    &\textcolor{gray}{fail}       &\textcolor{gray}{fail}        &\textcolor{gray}{fail}       &\textcolor{gray}{fail}       \\
                        &    & ImageNet-1k                &60.62       &94.18        &98.14        &58.80        &-       &-        &-        &-       &66.05    &96.59     &99.38     & \multicolumn{1}{c|}{72.98}    &-       &-       &-       &-       \\
                        &    & MMGL (Ours)                & \textbf{62.38}       &\textbf{95.39}        &\textbf{99.03}        &\textbf{60.91}        &\textbf{68.28}       &\textbf{97.48}        &\textbf{99.03}        &\textbf{53.67}       &\textbf{68.85}    &\textbf{97.20}     &\textbf{99.56}     & \multicolumn{1}{c|}{\textbf{74.54}}    &\textbf{76.47}       &\textbf{99.26}        &\textbf{99.40}       &\textbf{66.12} \\ \hline

\multirow{3}{*}{MPANet \cite{wu2021discover}}   &\multirow{3}{*}{CVPR 2021}        & \textcolor{gray}{Random Init.}               &\textcolor{gray}{fail}       &\textcolor{gray}{fail}        &\textcolor{gray}{fail}        &\textcolor{gray}{fail}        &\textcolor{gray}{fail}       &\textcolor{gray}{fail}        &\textcolor{gray}{fail}        &\textcolor{gray}{fail}       &\textcolor{gray}{fail}    &\textcolor{gray}{fail}     &\textcolor{gray}{fail}     & \multicolumn{1}{c|}{\textcolor{gray}{fail}}    &\textcolor{gray}{fail}       &\textcolor{gray}{fail}        &\textcolor{gray}{fail}       &\textcolor{gray}{fail}       \\
                        &    & ImageNet-1k                &70.58       &96.21        &98.80        &68.24        &75.58       &97.91        &\textbf{99.43}        &62.91       &76.74    &98.21     &\textbf{99.57}     & \multicolumn{1}{c|}{80.95}    &84.22       &99.66        &\textbf{99.96}       &75.11       \\
                        &    & MMGL (Ours)                & \textbf{72.45}       &\textbf{96.82}        &\textbf{98.90}        &\textbf{69.17}        &\textbf{76.43}       &\textbf{98.12}        &99.30        &\textbf{63.54}       &\textbf{78.24}    &\textbf{98.59}     &99.50     & \multicolumn{1}{c|}{\textbf{81.03}}    &\textbf{85.00}       &\textbf{99.68}        &\textbf{99.96}       &\textbf{75.75} \\
\end{tabular}
}
\end{table*}

\renewcommand\arraystretch{1}
\begin{table*}[t]\scriptsize
\centering
\caption{Cross-dataset performance evaluation on RegDB with Rank-1, 10, 20 (\%) and mAP (\%)  metrics.}
\label{Table:RegDB}
\begin{tabular}{c|c|c|c|cccc|cccc}
\multirow{2}{*}{Method}     & \multirow{2}{*}{Venue}      & \multirow{2}{*}{Pre-Train} & \multirow{2}{*}{Source} & \multicolumn{4}{c|}{Visble-Thermal}                                                        & \multicolumn{4}{c}{Thermal-Visble}                                                        \\
                            &                             &                            &                         & r1                   & r10                  & r20                  & mAP                   & r1                   & r10                  & r20                  & mAP                  \\ \Xhline{1pt}
\multirow{3}{*}{One-Stream \cite{ye2021deep}} & \multirow{3}{*}{TPAMI 2021} & \textcolor{gray}{Random Init.}                & \textcolor{gray}{RegDB}                   &\textcolor{gray}{17.04}                      &\textcolor{gray}{33.74}                      &\textcolor{gray}{44.76}                      &\textcolor{gray}{19.81}                       &\textcolor{gray}{16.70}                      &\textcolor{gray}{33.20}                      &\textcolor{gray}{44.37}                      &\textcolor{gray}{19.79}                      \\
                            &                             & ImageNet                & ImageNet-1k             & 63.17                     & 84.02                     & 91.89                     & 61.32                      & 61.39                     & 83.27                     & 90.99                     & 60.12                     \\
                            &                             & MMGL (Ours)                 & SYSU-MM01                    &\textbf{67.83}                      &\textbf{86.72}                      &\textbf{93.00}                      &\textbf{65.32}                       &\textbf{65.77}                      &\textbf{86.11}                      &\textbf{93.53}                      &\textbf{65.56}                      \\ \hline
\multirow{3}{*}{AGW \cite{ye2021deep}}        & \multirow{3}{*}{TPAMI 2021} & \textcolor{gray}{Random Init.}                & \textcolor{gray}{RegDB}                   &\textcolor{gray}{fail}                      &\textcolor{gray}{fail}                      &\textcolor{gray}{fail}                      &\textcolor{gray}{fail}                       &\textcolor{gray}{fail}                      &\textcolor{gray}{fail}                      &\textcolor{gray}{fail}                      &\textcolor{gray}{fail}                      \\
                            &                             & ImageNet                & ImageNet-1k             & 70.73                     & 86.46                     & 91.41                     & 65.04                      & 69.85                     & 86.31                     & 89.62                     & 63.66                     \\
                            &                             & MMGL (Ours)                 & SYSU-MM01                    &\textbf{75.43}                      &\textbf{89.91}                      &\textbf{94.02}                      &\textbf{70.12}                       &\textbf{74.25}                      &\textbf{88.60}                      &\textbf{91.05}                      &\textbf{70.00}                      \\ \hline
\multirow{3}{*}{DDAG \cite{ye2020dynamic}}       & \multirow{3}{*}{ECCV 2020}  & \textcolor{gray}{Random Init.}                & \textcolor{gray}{RegDB}                   &\textcolor{gray}{fail}                      &\textcolor{gray}{fail}                      &\textcolor{gray}{fail}                      &\textcolor{gray}{fail}                       &\textcolor{gray}{fail}                      &\textcolor{gray}{fail}                      &\textcolor{gray}{fail}                      &\textcolor{gray}{fail}                      \\
                            &                             & ImageNet                & ImageNet-1k             & \textbf{69.34}                     & \textbf{86.19}                     & \textbf{91.49}                     & \textbf{63.46}                     &  \textbf{68.06}                    &  \textbf{85.15}                    & \textbf{90.31}                     & \textbf{61.80}                     \\
                            &                             & MMGL (Ours)                 & SYSU-MM01                    & \multicolumn{1}{c}{67.28} & \multicolumn{1}{c}{85.63} & \multicolumn{1}{c}{89.95} & \multicolumn{1}{c|}{61.78} & \multicolumn{1}{c}{67.00} & \multicolumn{1}{c}{84.75} & \multicolumn{1}{c}{89.34} & \multicolumn{1}{c}{60.17} \\ \hline
\multirow{3}{*}{NFS \cite{chen2021neural}}        & \multirow{3}{*}{CVPR 2021}  & \textcolor{gray}{Random Init.}                & \textcolor{gray}{RegDB}                   & \multicolumn{1}{c}{\textcolor{gray}{37.92}} & \multicolumn{1}{c}{\textcolor{gray}{63.57}} & \multicolumn{1}{c}{\textcolor{gray}{72.83}} & \multicolumn{1}{c|}{\textcolor{gray}{38.05}} & \multicolumn{1}{c}{\textcolor{gray}{37.35}} & \multicolumn{1}{c}{\textcolor{gray}{62.81}} & \multicolumn{1}{c}{\textcolor{gray}{71.33}} & \multicolumn{1}{c}{\textcolor{gray}{37.69}} \\
                            &                             & ImageNet                & ImageNet-1k             & \multicolumn{1}{c}{80.54} & \multicolumn{1}{c}{91.96} & \multicolumn{1}{c}{95.07} & \multicolumn{1}{c|}{72.10} & \multicolumn{1}{c}{77.95} & \multicolumn{1}{c}{90.45} & \multicolumn{1}{c}{93.62} & \multicolumn{1}{c}{69.79} \\
                            &                             & MMGL (Ours)                 & SYSU-MM01                    & \multicolumn{1}{c}{\textbf{83.79}} & \multicolumn{1}{c}{\textbf{93.80}} & \multicolumn{1}{c}{\textbf{96.96}} & \multicolumn{1}{c|}{\textbf{76.38}} & \multicolumn{1}{c}{\textbf{81.72}} & \multicolumn{1}{c}{\textbf{93.23}} & \multicolumn{1}{c}{\textbf{96.25}} & \multicolumn{1}{c}{\textbf{75.08}}\\ \hline
\multirow{3}{*}{CAJ \cite{ye2021channel}}       & \multirow{3}{*}{ICCV 2021}  & \textcolor{gray}{Random Init.}                & \textcolor{gray}{RegDB}                   &\textcolor{gray}{fail}                      &\textcolor{gray}{fail}                      &\textcolor{gray}{fail}                      &\textcolor{gray}{fail}                       &\textcolor{gray}{fail}                      &\textcolor{gray}{fail}                      &\textcolor{gray}{fail}                      &\textcolor{gray}{fail}                      \\
                            &                             & ImageNet                & ImageNet-1k             & 83.25                     & 94.29                     & 97.04                     & 77.31                     &  82.63                    &  94.58                    & 97.11                     & 75.96                     \\
                            &                             & MMGL (Ours)                 & SYSU-MM01                    & \multicolumn{1}{c}{\textbf{84.32}} & \multicolumn{1}{c}{\textbf{95.27}} & \multicolumn{1}{c}{\textbf{97.65}} & \multicolumn{1}{c|}{\textbf{78.55}} & \multicolumn{1}{c}{\textbf{83.40}} & \multicolumn{1}{c}{\textbf{95.33}} & \multicolumn{1}{c}{\textbf{97.00}} & \multicolumn{1}{c}{\textbf{77.18}} \\ \hline
\multirow{3}{*}{MutualInfo \cite{tian2021farewell}}       & \multirow{3}{*}{CVPR 2021}  & \textcolor{gray}{Random Init.}                & \textcolor{gray}{RegDB}                   &\textcolor{gray}{fail}                      &\textcolor{gray}{fail}                      &\textcolor{gray}{fail}                      &\textcolor{gray}{fail}                       &\textcolor{gray}{fail}                      &\textcolor{gray}{fail}                      &\textcolor{gray}{fail}                      &\textcolor{gray}{fail}                      \\
                            &                             & ImageNet                & ImageNet-1k             & 73.2                     & -                     & -                     & 71.6                     &  71.8                    &  -                    & -                     & 70.1                     \\
                            &                             & MMGL (Ours)                 & SYSU-MM01                    & \multicolumn{1}{c}{\textbf{75.08}} & \multicolumn{1}{c}{\textbf{89.60}} & \multicolumn{1}{c}{\textbf{94.61}} & \multicolumn{1}{c|}{\textbf{72.74}} & \multicolumn{1}{c}{\textbf{75.30}} & \multicolumn{1}{c}{\textbf{89.21}} & \multicolumn{1}{c}{\textbf{91.17}} & \multicolumn{1}{c}{\textbf{74.83}} \\ \hline
                            
\multirow{3}{*}{MPANet \cite{wu2021discover}}       & \multirow{3}{*}{CVPR 2021}  & \textcolor{gray}{Random Init.}                & \textcolor{gray}{RegDB}                   &\textcolor{gray}{fail}                      &\textcolor{gray}{fail}                      &\textcolor{gray}{fail}                      &\textcolor{gray}{fail}                       &\textcolor{gray}{fail}                      &\textcolor{gray}{fail}                      &\textcolor{gray}{fail}                      &\textcolor{gray}{fail}                      \\
                            &                             & ImageNet                & ImageNet-1k             & 82.8                     & -                     & -                     & 80.7                     &  83.7                    &  -                    & -                     & 80.9                     \\
                            &                             & MMGL (Ours)                 & SYSU-MM01                    & \multicolumn{1}{c}{\textbf{82.95}} & \multicolumn{1}{c}{\textbf{94.26}} & \multicolumn{1}{c}{\textbf{97.15}} & \multicolumn{1}{c|}{\textbf{81.23}} & \multicolumn{1}{c}{\textbf{85.28}} & \multicolumn{1}{c}{\textbf{96.34}} & \multicolumn{1}{c}{\textbf{98.00}} & \multicolumn{1}{c}{\textbf{83.89}}
\end{tabular}
\end{table*}

\subsection{RGB-IR ReID with MMGL}
We first use the proposed method to pre-train seven state-of-the-art ReID models, including One-Stream \cite{ye2021deep}, AGW \cite{ye2021deep}, DDAG \cite{ye2020dynamic}, NFS \cite{chen2021neural}, CAJ \cite{ye2021channel}, MultualInfo \cite{tian2021farewell}, and MPANet \cite{wu2021discover}. Firstly, a random initialized backbone is pre-trained with MMGL on SYSU-MM01. Then we fine-tune the pre-trained checkpoint with default settings to perform supervised RGB-IR ReID.

As shown in Table \ref{Table:SYSU}, when straightforwardly performing supervised training from scratch on SYSU-MM01, much inferior accuracy is observed for all methods. DDAG, CAJ, MutualInfo, and MPANet even encounter gradient explosion and fail to converge. The exploding gradients mainly comes from the triplet loss in Eq.(\ref{loss:ReID}), which suggests that random initialized models cannot form a semantically meaningful Euclidean feature space. 

Our proposed MMGL pre-training strategy achieves consistent and significant improvement (more than \textbf{25\%} Rank-1 and mAP boost in average) over baselines without pre-training. Such performance gains are obtained neither using additional training data nor by supervised pre-training. This manifests that the proposed self-learning technique and part-aware cycle-contrastive constraint provide data-efficient and powerful regularization to prevent over-fitting.

More strikingly, MMGL pre-trained model even outperforms its ImageNet-supervised counterpart on most state-of-the-arts. Particularly, on the AGW baseline, MMGL achieves 8.03\% absolute Rank-1 improvement to ImageNet pre-training. This indicates that elimination of modality bias has huge positive effects on cross-modality ReID. Furthermore, the promising results on NFS, DDAG, MutualInfo, and MPANet demonstrate that MMGL is robust to different backbones and loss functions.

Another noteworthy feature of our approach is its fast convergence speed. As shown in Fig. \ref{fig:problem}, when pre-training on SYSU-MM01 with AGW baseline, MMGL has vastly reduced the amount of training time from seven days to four hours, while without any time increase in fine-tuning. This results in more efficient RGB-IR ReID.

\subsection{Generalization Ability across Datasets}
We further evaluate the cross-dataset generalization capability of MMGL features. Specifically, we pre-train the aforementioned seven models with MMGL on SYSU-MM01, and then transfer the learned features to solve RGB-IR ReID on RegDB. For fair comparison, we use the same fine-tuning hyperparameters as the ImageNet-pretrained model has, and provide random initialization results as references.

Table \ref{Table:RegDB} shows the fine-tuning results on RegDB. We can see that the performance gap between no pre-training and ImageNet pre-training widens more than that on SYSU-MM01, possibly owing to the even smaller size of RegDB. Random initialization with scarce data also makes it hard to train models from scratch --- AGW, DDAG, and CAJ, MutualInfo, and MPANet all meet gradient explosion on RegDB. Moreover, light-weight models, i.e., One-Stream and NFS, seem robust to the initialization strategies. However, they still suffer from serve over-fitting and achieve much inferior ReID accuracy due to the data scarcity.

Surprisingly, MMGL pre-training demonstrates consistent performance and good transferability across datasets, and even surpasses its ImageNet-supervised counterpart when outfitted on various off-the-shelf models. Note that such improvement is achieved along with large domain shift, as SYSU-MM01 is captured by near-infrared cameras while RegDB is collected by far-infrared sensors. All of the above results suggest that MMGL features are robust and generalizable.

\subsection{Comparisons with Other SSL Methods}

We compare MMGL with eight self-supervised learning approaches on the SYSU-MM01 dataset, including relevant baseline and state-of-the-art algorithms. The involved methods can be roughly divided into two categories: \textbf{1)} pretext task design (Colorization \cite{zhang2016colorful}, Jigsaw \cite{noroozi2016unsupervised}, PSL \cite{li2021progressive}) and \textbf{2)} contrastive learning (SimCLR \cite{chen2020simple}, SwAV \cite{caron2020unsupervised}, BYOL \cite{grill2020bootstrap}, SimSiam \cite{chen2021exploring}, and Barlow Twins \cite{zbontar2021barlow}).

\linespread{1.2}
\begin{table}[h]
\caption{Comparison with other SSL methods on SYSU-MM01 (Single-Shot \& All-Search).}
\centering
\label{Table:SSL}
\begin{tabular}{l|ccc}
Method     & r1 & r10  & mAP \\ \Xhline{1.5pt}
Colorization \cite{zhang2016colorful}     & 30.57    & 69.02          & 29.49    \\
Jigsaw \cite{noroozi2016unsupervised}   & 40.74   & 82.07         & 37.71    \\ 
PSL \cite{li2021progressive}   & 50.86   & 86.12         & 49.35    \\ \hline
SimCLR \cite{chen2020simple}     & 46.17    & 84.93    &  43.38        \\ 
SwAV \cite{caron2020unsupervised}     & 48.25  & 85.62     & 45.07        \\ 
BYOL \cite{grill2020bootstrap}     & 52.37    & 87.34    &  50.03        \\ 
SimSiam \cite{chen2021exploring}     & \textcolor{gray}{fail}    & \textcolor{gray}{fail}    &  \textcolor{gray}{fail}        \\
Barlow Twins \cite{zbontar2021barlow}     & 51.78    & 86.45    &  50.23        \\ \hline
MMGL (Ours) &\textbf{56.61}    &\textbf{90.23}         &\textbf{54.50}    
\end{tabular}
\end{table}
\linespread{1}

From the results listed in Table \ref{Table:SSL}, we can see that Colorization produces modest accuracy on the RGB-IR ReID task. This is because IR images inherently contain no color information. Jigsaw and PSL are more complex and most similar pretext tasks to permutation recovery, but counter-intuitively, both yielding suboptimal performance. One possible reason is that the pose variations in person images raise severe spatial misalignment in horizontal direction, rendering them less effective in learning discriminative features than MMGL. 

Contrastive learning (CL) methods share similar learning objectives with our proposed PCC constraint, namely instance discrimination \cite{he2020momentum}. They perform better than other pretext tasks, yet still attain inferior results than MMGL. This is because existing CL methods heavily rely on intra-modality data augmentation to create multiple views of the same sample, which does not explicitly consider the modality discrepancy in the task. We see that SimSiam even fails to converge on SYSU-MM01. This suggests that the modality gap further aggravates the mode collapse issue of SSL \cite{chen2021exploring}, which cannot be effectively alleviated by simple stop-gradient operations. Instead, Our PCC utilizes cross-modality cycle-consistency to generate positive patch pairs, effectively eliminating such issue and thus leading to more modality-invariant representations.

Notably, it has been proven that existing CL methods do not work well with small batch sizes \cite{chen2020simple}. However, person ReID is a typical single-GPU task where the efficiency greatly matters \cite{ye2021deep}. Our proposed paradigm brilliantly exploits body partitions to increase the amount of negative samples with negligible memory costs. Generally, MMGL outperforms state-of-the-art SSL methods consistently.

\subsection{Ablation Studies}
We evaluate the effectiveness of each MMGL component on the SYSU-MM01 dataset in both \textit{all search} and \textit{indoor search} modes. For fair comparison, all ablations are conducted on AGW \cite{ye2021deep} with fixed default hyper-parameters. Specifically, $\mathcal{P}$ denotes permutation recovery, $\mathcal{C}$ represents the PCC constraint in Eq. \ref{eq:PCC}, $\mathcal{E}$ represents the penalization term in Eq. \ref{eq:all_loss}.

\renewcommand\arraystretch{1.2}
\begin{table}[h]
\linespread{2}
\caption{Ablation Studies on the SYSU-MM01 Dataset (Single-Shot).}
\vspace{-0.5cm}
\label{Table:ablation}
\begin{center}

\begin{tabular}{l|ccc|ccc}
\multirow{2}{*}{Method} & \multicolumn{3}{c|}{All-Search} & \multicolumn{3}{c}{Indoor-Search} \\ 
                        & r1        & r10      & mAP      & r1        & r10       & mAP       \\ \Xhline{1.5pt}
\textcolor{gray}{Rand Init}                     & \textcolor{gray}{25.47}          & \textcolor{gray}{66.06}         & \textcolor{gray}{26.07}         & \textcolor{gray}{26.38}          & \textcolor{gray}{70.61}          & \textcolor{gray}{35.95}          \\
ImageNet-1k                       & 48.58     & 87.58    & 49.37    & 54.73     & 92.46     & 63.72     \\
Huang \textit{et al.} \cite{huang2021alleviating}                      & 49.0          & -         & 48.8         & -           & -          & -          \\
$\mathcal{P}$                     & 50.99          & 88.25         & 49.77         & 55.53           & 91.29          & 63.33          \\
$\mathcal{P}+\mathcal{E}$                  & 55.57           & 90.67         & 53.99         & 60.32           & 94.41          & 67.81          \\
$\mathcal{P}+\mathcal{C}$                  & 55.05           & 89.61         & 53.05         & 58.93           & 93.23          & 66.34          \\
$\mathcal{P}+\mathcal{C}+\mathcal{E}$                  & \textbf{56.61}           & \textbf{90.23}         & \textbf{54.50}         & \textbf{62.55}           & \textbf{94.60}          & \textbf{69.62}          \\
\end{tabular}
\end{center}
\end{table}



As shown in Table \ref{Table:ablation}, when only pre-training with the $\mathcal{P}$ task, AGW presents encouraging performance even better than ImageNet pre-training. One possible explanation is that permutation recovery task explicitly exploits body typology, a natural attribute being invariant to both modalities, as pre-training supervision. With this supervision, each cross-modality image pair is mapped into a shared permutation latent space where modality distributions can be better-aligned. It is noteworthy that this result is achieved without extra data and labels, suggesting that supervised pre-training on large RGB image sets does not bring much improvement to downstream cross-modality tasks. We can see that $\mathcal{P}$ also outperforms Huang \textit{et al.} \cite{huang2021alleviating}, which aims to alleviate modality bias training via generating third modality person images for fine-tuning. This indicates that, compared with image generation, multi-modal pre-training serves as a more essential and effective solution to the modality bias training issue.

The performance boost by sole permutation recovery is modest, as the model may easily find undesired trivial solutions, \textit{e.g.,} by directly exploiting boundary patterns of patches to solve the task. But, when imposing the proposed PCC constraint on patch embeddings ($\mathcal{P} + \mathcal{C}$), significant improvement can be observed (\textbf{+4.06\%} Rank-1, \textbf{+3.28\%} mAP). Such improvement mainly comes from two aspects. On the one hand, PCC offers discriminative features via instance discrimination. On the other hand, it ensures modality invariance by attracting cycle-retrieved RGB-IR pairs.

Consider that the network is initialized with random weights, PCC may retrieve random positive pairs in early training and undermine the semantic structure of the learned feature space. We fix this issue by introducing the penalization term $\mathcal{E}$, which aims at classifying each patch into its corresponding position label. As can be seen in the 7th rows of Table \ref{Table:ablation}, $\mathcal{E}$ further improves the Rank-1 accuracy by 1.56\%. This demonstrates that patch-wise classification can effectively guide $\mathcal{P}$ and $\mathcal{C}$ to discover more semantically consistent patch pairs. Notably, directly imposing $\mathcal{E}$ on $\mathcal{P}$ also leads to 4.58\% Rank-1 improvement, which verifies its capacity in learning discriminative local representations.

\textbf{Hyper-Parameter Analysis.} Here, we analyse the influence of three hyper-parameters of MMGL: the number of partition stripes $N$, the temperature $\tau$ (Eq. (\ref{eq:PCC})), and the balance coefficient $\lambda$ (Eq. (\ref{eq:all_loss})). All the experiments are conducted with the AGW baseline \cite{ye2021deep}.

Since the size of \textit{res5} feature map in this work is $18\times9\times2048$, we choose $N$ from the set of $\{2, 3, 6, 9, 18\}$. As listed in Table \ref{Table:N}, the identification accuracy gradually improved as $N$ increases. The results suggests that smaller $N$ makes the model much easier to find shortcut solutions that severely degrades the quality of learned representations. However, larger $N$ does not mean better accuracy. As we can see, there is a significant performance drop with $N=18$. This is because, as $N$ grows, the complexity of the task also increases. The model has to seek more high-level semantics to solve the task, which makes it even more harder to converge during training. Empirically, We find that $N=6$ is a good trade-off between complexity and accuracy. 
\vspace{-0.3cm}
\begin{table}[h]\small
\centering
\caption{Results on SYSU-MM01 (Single-Shot \& All-Search) with different partition stripe numbers $N$. }
\label{Table:N}
\begin{tabular}{c|ccccc}
$N$      & 2 & 3 & 6 & 9 & 18 \\ \Xhline{1.5pt}
Rank-1 &52.54   &55.09   &\textbf{56.61}   & 55.70  &\textcolor{gray}{fail}    \\
mAP    &50.77   &53.08   &\textbf{54.50}   & 53.16  &\textcolor{gray}{fail}  
\end{tabular}
\end{table}

The temperature $\tau$ serves as a core hyper-parameter of contrastive learning that controls the penalty strength on hard negative samples \cite{wang2021understanding}. We tunes the temperature following \cite{he2020momentum} and the results summarized in Table \ref{Table:tau} show that $\tau=$ 0.07 is a good choice for the downstream RGB-IR ReID task.
\vspace{-0.3cm}
\begin{table}[h]\small
\centering
\caption{Results on SYSU-MM01 (Single-Shot \& All-Search) with different temperature $\tau$. }
\label{Table:tau}
\begin{tabular}{c|cccccc}
$\tau$       & 0.03 & 0.05 & 0.07 & 0.1 & 0.2 & 0.3 \\ \Xhline{1.5pt}
Rank-1 &52.39      &54.11      &\textbf{56.61}      &55.85     &54.79     &54.03     \\
mAP    &51.21      &53.47      &\textbf{54.50}      &53.97     &53.91     &53.00    
\end{tabular}
\end{table}

$\lambda$ is a coefficient to balance each learning objective of MMGL. As shown in Table \ref{Table:l}, with different $\lambda$, our MMGL consistently surpasses ImageNet pre-training (48.58\% Rank-1 \& 49.37\% mAP) by a large margin. And $\lambda=0.2$ is a good trade-off to balance each objective.
\vspace{-0.3cm}
\begin{table}[h]\small
\centering
\caption{Results on SYSU-MM01 (Single-Shot \& All-Search) with different balance coefficient $\lambda$.}
\label{Table:l}
\begin{tabular}{c|cccccc}
$\lambda$       & 0.1 & 0.2 & 0.3 & 0.4 & 0.5 & 1 \\ \Xhline{1.5pt}
Rank-1 &55.05      &\textbf{56.61}      &55.78      &54.75     &55.88     &54.03     \\
mAP    &53.15      &\textbf{54.50}      &53.80      &53.00     &53.56     &53.11    
\end{tabular}
\end{table}

\subsection{Visualization Results}

\begin{figure}[t]
\begin{center}
   \includegraphics[width=1\linewidth]{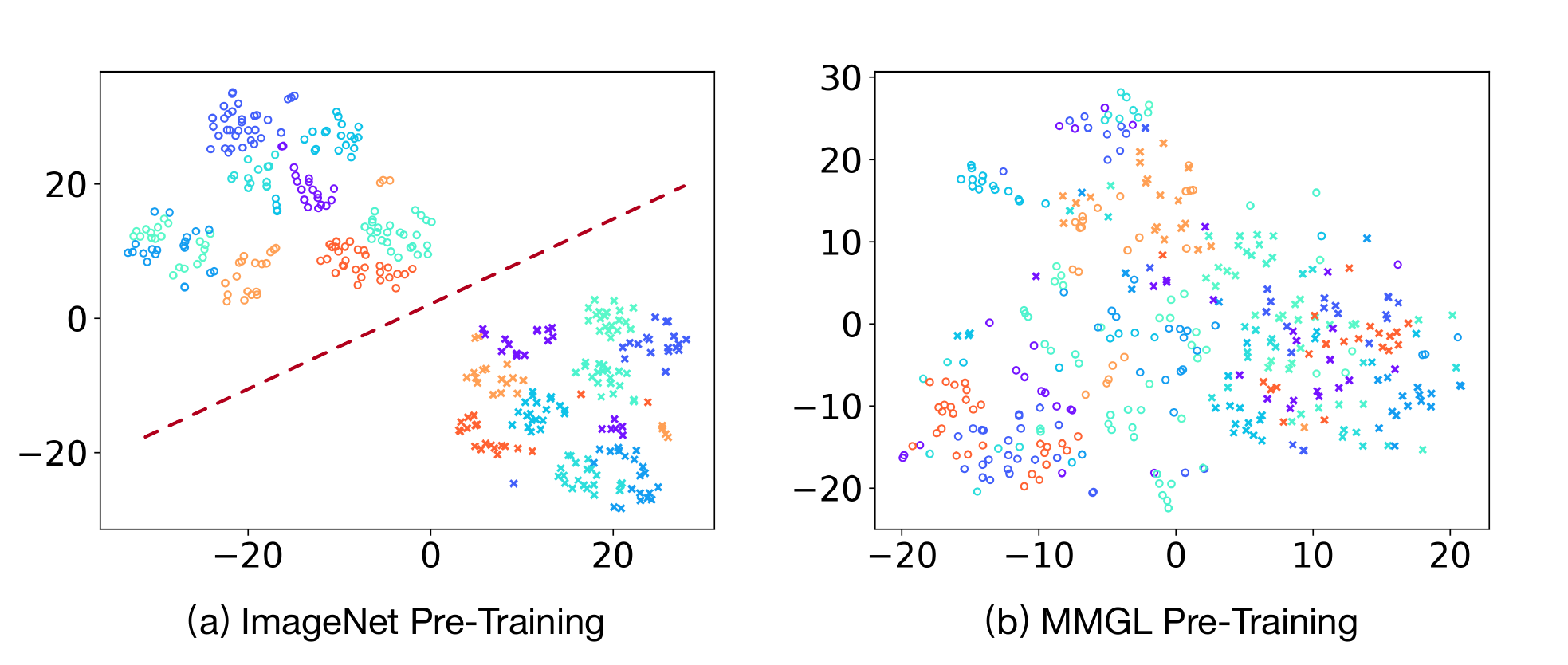}
\end{center}
   \caption{Visualization of the pre-learned features. Different colors represent distinct identities, while circle and cross symbols denote RGB and IR modalities, respectively. The experiment is conducted on SYSU-MM01 dataset with the AGW baseline \cite{ye2021deep}.}
\label{fig:tsne_pretrain}
\end{figure}

\textbf{Visualization of the Pre-Learned Features.} To better understand the proposed pre-training paradigm, on the SYSU-MM01 dataset, with AGW \cite{ye2021deep}, we visualize the \textit{t}-SNE distribution \cite{van2008visualizing} of features pre-learned by ImageNet pre-training and our MMGL approach. As shown in Fig. \ref{fig:tsne_pretrain}, we randomly sampled 10 identities from the SYSU-MM01 test set and used all corresponding images for visualization. Observations are summarized as follows.

\begin{figure}[t]
\begin{center}
   \includegraphics[width=1\linewidth]{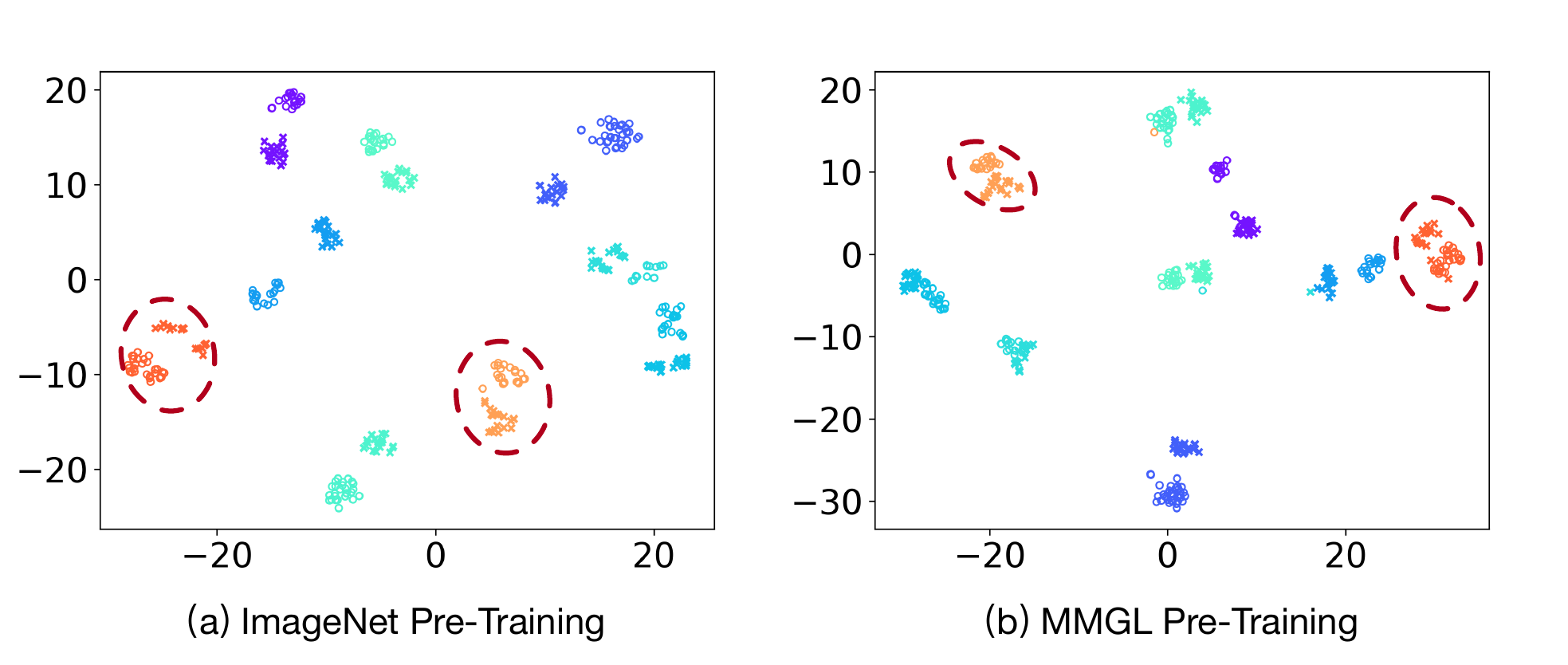}
\end{center}
   \caption{Visualization of the fine-tuned features. Different colors represent distinct identities, while circle and cross symbols denote RGB and IR modalities, respectively. The experiment is based on SYSU-MM01 dataset with the AGW baseline \cite{ye2021deep}.}
\label{fig:tsne_finetune}
\vspace{-0.3cm}
\end{figure}

With ImageNet pre-training, there exists huge distribution gap between the RGB and IR modalities (as shown by the red line of Fig. \ref{fig:tsne_pretrain} (a)). This is because ImageNet pre-training is essentially a single-modality paradigm, and the pre-trained backbones only have one stream. At the fine-tuning stage, we have to use shared weights to initialize independent blocks of the two-stream AGW baseline, leading to \textit{modality bias training}. Besides, although ImageNet pre-training can yield features with promising intra-class compactness, such an advantage mainly comes from large-scale \textbf{labeled} data, which are highly expensive.

Fig. \ref{fig:tsne_pretrain} (b) presents the feature distribution of our MMGL pre-training. One inspiring observation is that the modality gap has been almost eliminated. The underlying reasons are two-fold. On the one hand, MMGL allows us to pre-train a two-stream backbone directly on target multi-modal datasets, which can achieve a balance between RGB and IR modalities. On the other hand, the permutation recovery pretext task explicitly forms a modality-shared latent permutation space to further narrow the modality gap. 

Finally, since MMGL is self-supervised, it may present inferior intra-class compactness than ImageNet supervised pre-training. We trade some intra-class compactness for better modality invariance because the latter is more important for pre-training cross-modality ReID models \cite{ye2020visible}. Empirically, such a trade-off significantly improves the pre-training efficiency and even the final performance. To summarize, MMGL features are modality-invariant, cost-effective, and powerful.

\textbf{Visualization of the Fine-Tuned Features.} We further visualize the fine-tuned features of ImageNet and MMGL pre-training under the same settings as Fig. \ref{fig:tsne_pretrain}. As can be seen in Fig. \ref{fig:tsne_finetune}, after supervised fine-tuning, the modality gap between MMGL features is still more narrowed than the ImageNet counterparts (see the red ellipse regions). This indicates that the pre-learned modality invariance is transferable to the downstream RGB-IR ReID task. Interestingly, different from the pre-training stage, MMGL exhibits better intra-class compactness and inter-class discriminability than ImageNet pre-training, which verifies that the modality invariance is more important than discriminability in cross-modality settings.

\begin{figure*}[t]
\begin{center}
   \includegraphics[width=1\linewidth]{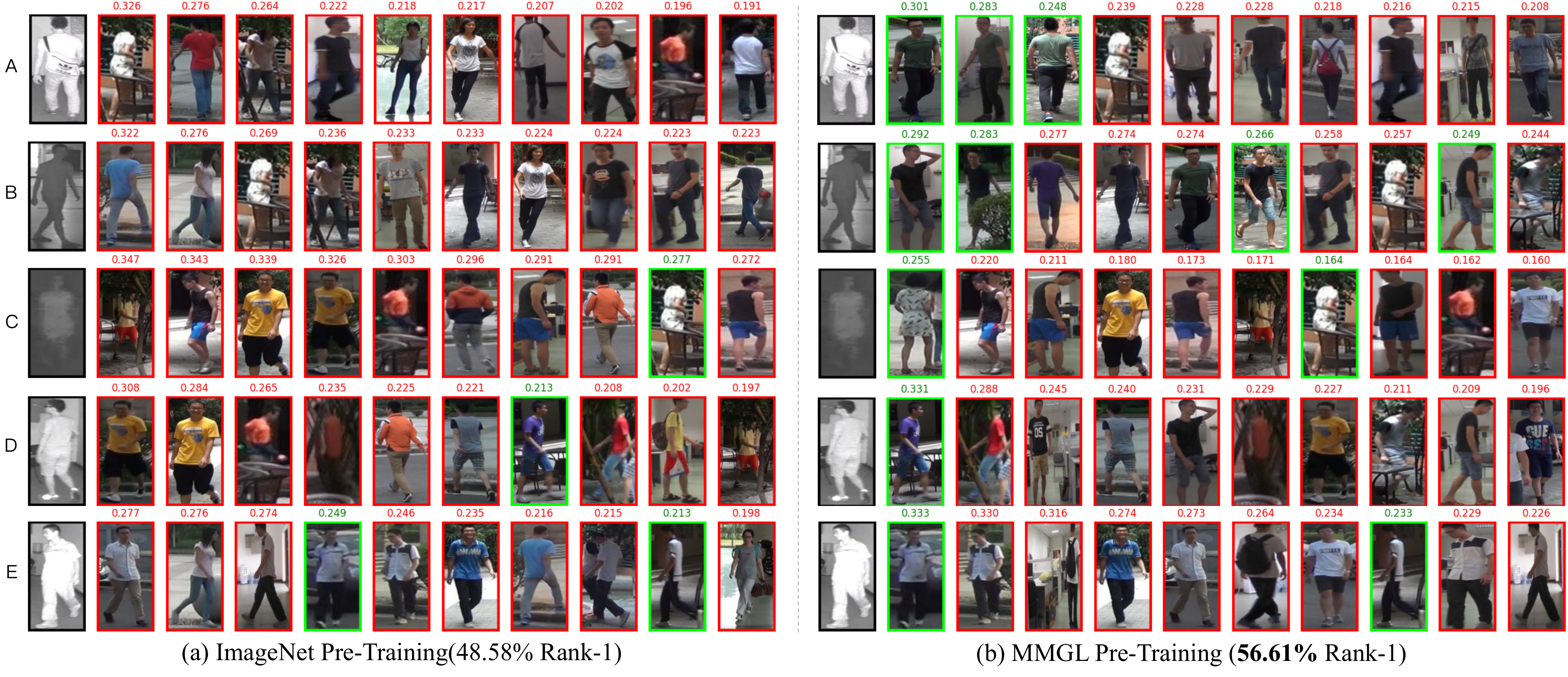}
\end{center}
   \caption{Comparison of the top-10 retrieved results of ImageNet and our MMGL pre-training on the SYSU-MM01 dataset, AGW baseline \cite{ye2021deep}. The green bounding boxes indicate the correct matchings and red ones represent wrong matchings (Best viewed in color).}
\label{fig:retrieval}
\end{figure*}

\textbf{Visualization of the Retrieval Results.} We compare the top-10 retrieved results of 5 randomly selected query samples between our MMGL and ImageNet pre-training, respectively (on SYSU-MM01 with AGW baseline). In Fig. \ref{fig:retrieval}(a) and (b), the first column includes the query samples, and retrieval results are ranked from left to right according to descending cosine similarity scores. 

As we can see, there are plenty of wrong matchings with ImageNet pre-training, attaining 48.58\% Rank-1 accuracy. Instead, our MMGL pre-training returns right matchings with a 56.61\% Rank-1 even though some cases are hard for human to recognize (e.g., query A and C). This demonstrates the effectiveness of MMGL in boosting feature discriminability. Interestingly, we discover that, in some cases (e.g., query A), even if people change their belongings, MMGL pre-training is still capable of achieving accurate image retrieval by mining other visual cues, perhaps the body shape.

\begin{figure}[h]
\begin{center}
   \includegraphics[width=1\linewidth]{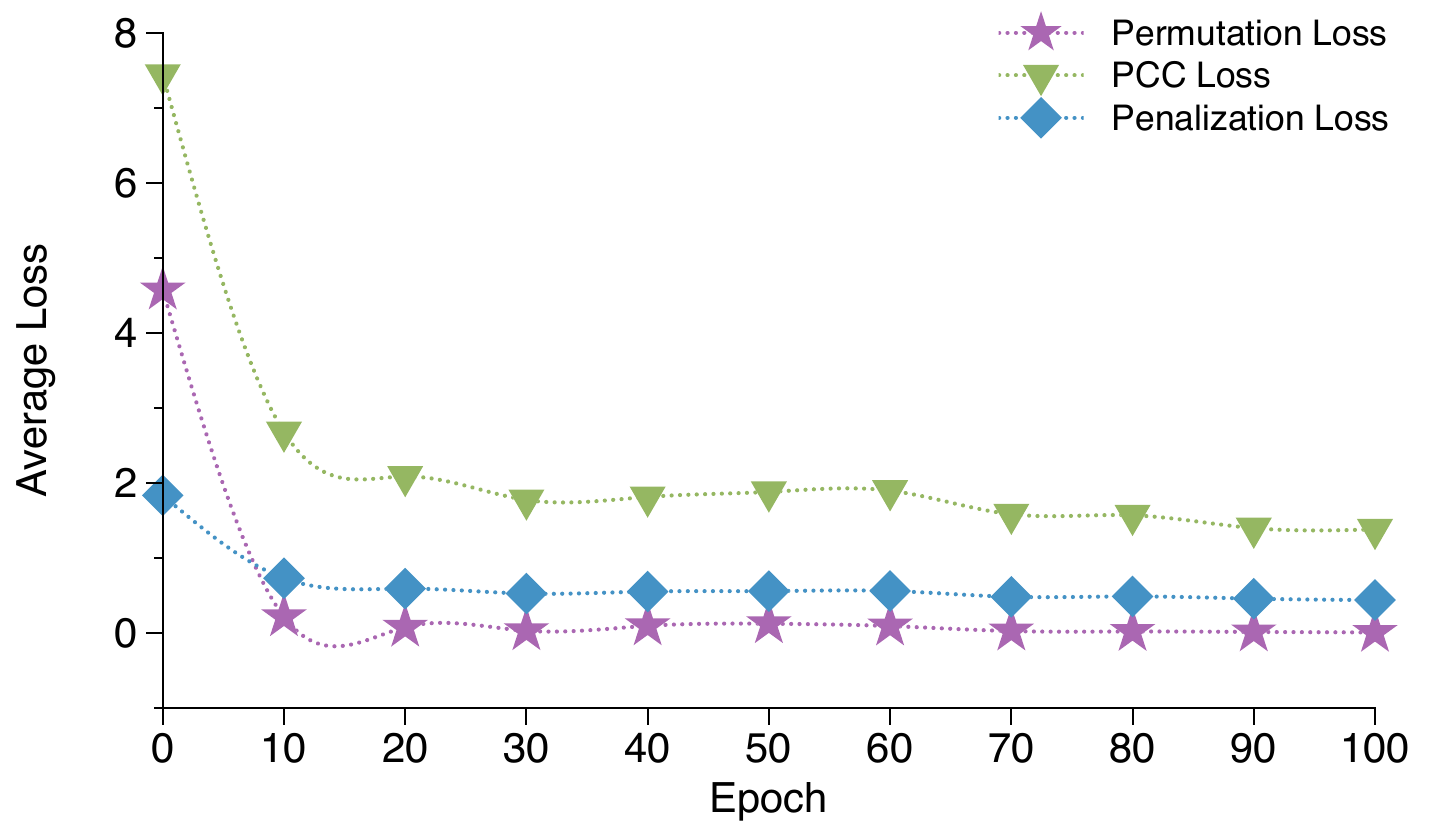}
\end{center}
   \caption{Training curve of MMGL pre-training on AGW \cite{ye2021deep}, SYSU-MM01. Obviously, MMGL has fast converge speed and can coverage in only 40 epochs. Here the coefficient of PCC loss is 0.2, while coefficients of the other losses is set to 1. PCC = Part-aware cycle-contrastive learning.}
\label{fig:loss}
\end{figure}

\textbf{Visualization of the Pre-Training Converge Speed.} A noteworthy feature of MMGL is its fast converge speed. Fig. \ref{fig:loss} illustrates the training curve of MMGL with the AGW baseline \cite{ye2021deep} on SYSU-MM01 dataset. The total 100-epochs training ends only about 6 hours on a single Nvidia RTX 2080Ti GPU. Compared with the expensive ImageNet supervised pre-training, our proposed method significantly boosts the training efficiency, while simultaneously ensuring competitive ReID accuracy.

\begin{figure}[h]
\begin{center}
   \includegraphics[width=1\linewidth]{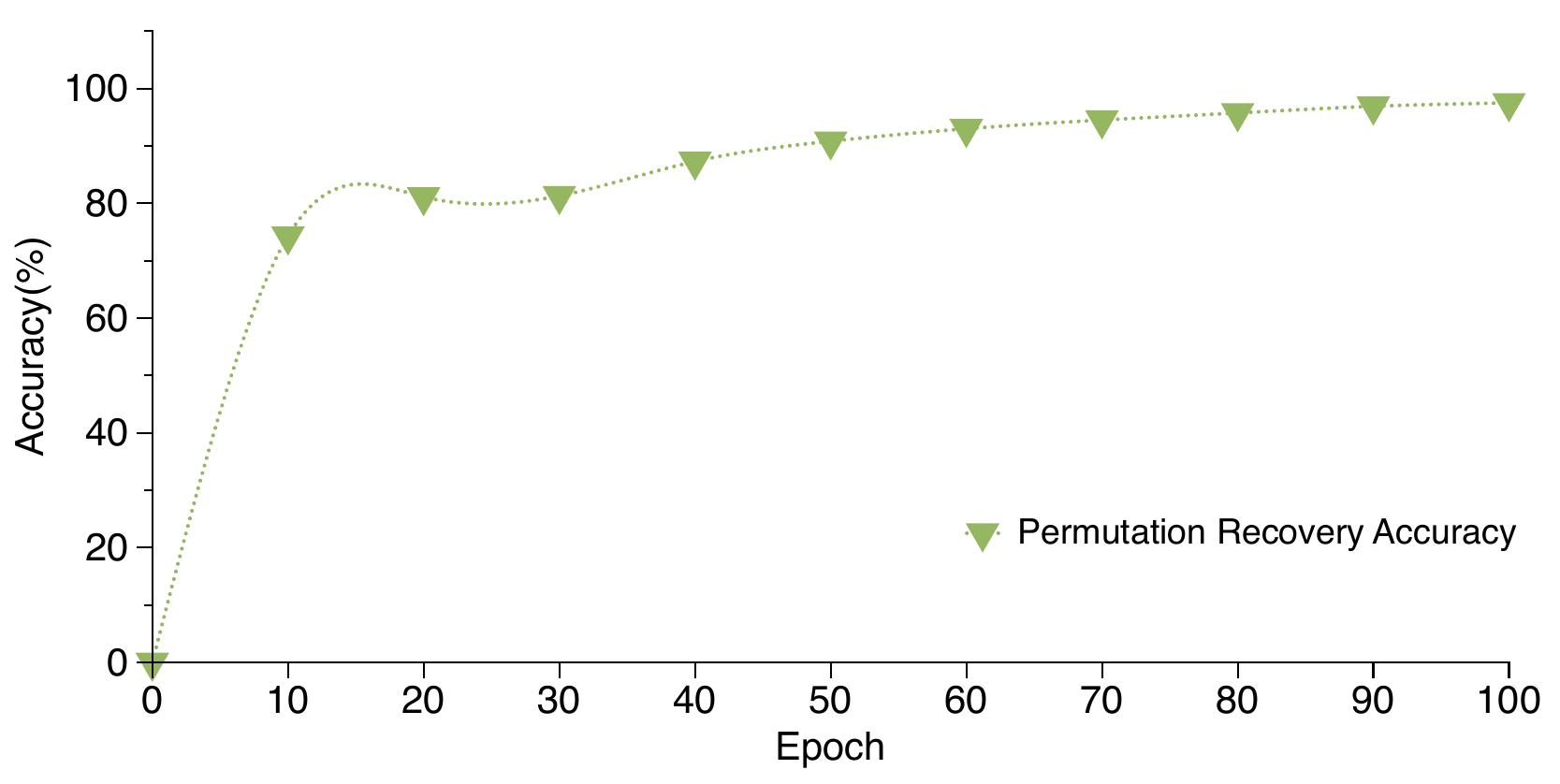}
\end{center}
\vspace{-0.3cm}
   \caption{Permutation recovery accuracy of MMGL on the official SYSU-MM01 validation set \cite{wu2017rgb}. The optimal accuracy is 97.5\%.}
\label{fig:permu_acc}
\end{figure}

\textbf{Visualization of the Permutation Recovery Accuracy.} Another notable feature of MMGL is that we can directly evaluate the quality of pre-learned representations using the permutation recovery accuracy. This allows us to select the best model based on an validation set. As shown in Fig. \ref{fig:permu_acc}, MMGL achieves 97.5\% permutation recovery accuracy on the SYSU-MM01 validation set \cite{wu2017rgb}, which suggests that permutation recovery serves as a good complexity-performance trade-off for pre-training RGB-IR ReID models.

\section{Discussions}
\label{sec:discussion}
To our best knowledge, MMGL is the first work on pre-training solutions for RGB-IR ReID. We find that directly adopting self-supervised pre-training on target multi-modal datasets yields competitive results against ImageNet supervised pre-training. We discuss the broader impact for pre-training solutions from three aspects.

Firstly, without ImageNet pre-training, the performance gains brought by some SOTAs could no longer exist. For example, as shown in Table \ref{Table:SYSU}, pre-trained with ImageNet, nearly 6.17\% absolute performance improvement is brought by DDAG compared with AGW. But surprisingly, AGW outperforms DDAG by a large margin when training both from scratch. This motivates us to rethink whether it's just because recent SOTAs cater better for ImageNet pre-trained checkpoints that they play a role in RGB-IR ReID. That is, existing methods work probably only because they could effectively alleviate the modality bias caused by ImageNet pre-training. But as for whether these approaches themselves are robust to multi-modal scenarios is still unclear. Since MMGL can essentially eliminates the modality bias in the pre-training stage, we believe it can serve as a `\textit{Touchstone}' to evaluate the robustness of existing cross-modality ReID methods. We leave it to future work.

Secondly, recent contrastive learning methods often rely heavily on larger batch size/memory bank \cite{chen2020simple} or a momentum encoder \cite{he2020momentum} to improve the quality of pre-learned representations, which is computationally intensive. Unlike these approaches, MMGL is simple, easy-to-implement and works well with only batch size of 96, a siamese network, and the single-GPU setting. That makes it better suitable for person ReID tasks in which the efficiency greatly matters.  

To our understanding, the key point of designing self-supervised pre-training paradigms for RGB-IR ReID lies in how to mine cross-modal correspondence without manual labels. In this paper, we have shown that learning with shared permutation labels and cycle-consistent cross-modality patch pairs can alleviate the modality gap in a self-supervised manner, leading to significant performance boosts to existing RGB-IR ReID methods. It will be very exciting to witness more and more interesting ideas. Besides, several studies also suggest that self-supervised learning could benefit more from large model sizes and data scale than supervised learning \cite{chen2020simple,he2020momentum,grill2020bootstrap}. However, to our best knowledge, there does not exist an ImageNet-scale benchmark for cross-modality ReID. Currently the largest RGB-IR ReID dataset, SYSU-MM01 \cite{wu2017rgb}, contains about 50,000 images, which is still much smaller than ImageNet. Although MMGL is self-supervised and can avoid labor intensive manual annotation, data still matter!

\begin{figure}[h]
\begin{center}
   \includegraphics[width=1\linewidth]{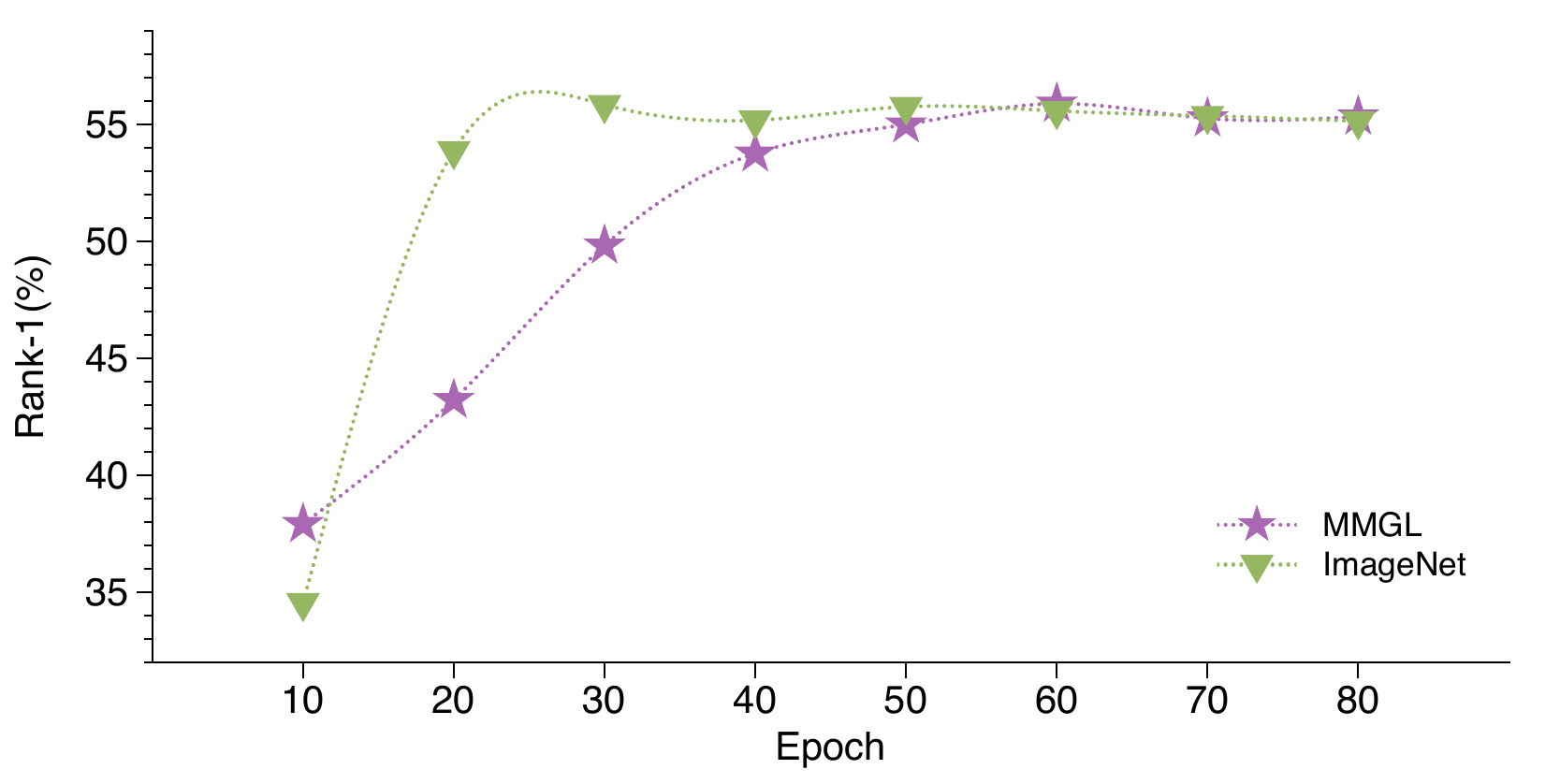}
\end{center}
\vspace{-0.5cm}
   \caption{Comparison between the Rank-1 accuracy variation curves of MMGL and ImageNet pretraining on SYSU-MM01 validation set \cite{wu2017rgb} (Single-Shot \& All-Search). The experiment is based on DDAG \cite{ye2020dynamic}.}
\label{fig:finetune}
\end{figure}

There are also several limitations of our method. First, although the pre-training speed is fast (see Fig. \ref{fig:problem}), MMGL pre-trained models converge slower than their ImageNet pre-trained counterparts in the fine-tuning stage. Fig. \ref{fig:finetune} illustrates DDAG's \cite{ye2020dynamic} Rank-1 accuracy variation curve on the SYSU-MM01 validation set, initialized with ImageNet (green line) and MMGL (purple line) checkpoints, respectively. Obviously, ImageNet pre-training endows DDAG faster converge speed than MMGL. One possible reason is that ImageNet is of a much larger order of magnitude than our pre-training dataset SYSU-MM01, so that it can offer more accurate batch statistics and stable gradient flow in early training \cite{jing2020self}.

Another limitation of MMGL is its varying performance boost on existing state-of-the-arts. For example, MMGL brings a 8.03\% absolute improvement in Rank-1 accuracy on AGW \cite{ye2021deep}, while the performance boost on DDAG \cite{ye2020dynamic} is modest ($<$1\% absolute improvement in Rank-1 accuracy). 

We argue that such a limitation is predictable for two reasons. Firstly, In experiments we keep the default settings (e.g., hyper-parameters and data augmentation strategies) same as the original implementation of each method. These settings are all carefully customized for ImageNet pre-training, which might NOT be the best choices for MMGL. Secondly, current SOTA models often present distinct parameter complexities. For instance, DDAG introduces complex attention modules (\textbf{$\sim$2M} extra parameters) for feature enhancement, which means it is \textit{data-hungery}. It is pretty hard to provide a good start point for such a model using only \textbf{0.05M} pre-training images.

\section{Conclusion}
\label{sec:conclusion}
This paper makes the first attempt to investigate the pre-training solution for RGB-IR cross-modality person ReID. To overcome the modality bias issue brought by ImageNet pre-training, we propose MMGL, a self-supervised pre-training paradigm that allows ReID models to be pre-trained and fine-tuned directly on existing cross-modality pedestrian datasets, showing superior performance and robustness against over-fitting. By solving a permutation recovery pretext task, MMGL learns highly-invariant representations across modalities. We further propose a part-aware cycle-contrastive learning strategy to learn correspondence between unpaired RGB-IR patches, significantly improving the discriminability of local features. Extensive experiments reveal the effectiveness and transferability of MMGL, even surpassing its ImageNet supervised counterpart without extra data or manual labels.

\bibliographystyle{IEEEtran}
\bibliography{ref}

\vfill

\end{document}